\documentclass[review]{elsarticle}
\usepackage[whole]{bxcjkjatype}

\usepackage{graphicx}
\usepackage{floatrow}

\usepackage[font=footnotesize]{subcaption}
\usepackage[font=footnotesize]{caption}

\usepackage{stfloats}
\usepackage {diagbox}

\usepackage[ruled,linesnumbered]{algorithm2e}
\SetAlgorithmName{Alg.}{Alg.}{List of Algorithms}

\usepackage{amssymb, amsmath, mathrsfs, times, cancel} % additional math functions
\newcommand{\argmin}{\mathop{\rm arg~min}\limits} % math form arg min
\newcommand{\indep}{\mathop{\perp\!\!\!\!\perp}}

\usepackage{url}

\usepackage[inline]{enumitem}\setlist[enumerate]*{label=(\roman*)}

\newcommand\figref[1]{\text{Fig.~\ref{#1}}}
\newcommand\tableref[1]{\text{Table ~\ref{#1}}}
\newcommand\equref[1]{\text{Eq.~(\ref{#1})}}
\newcommand\algorithmref[1]{\text{Alg.~\ref{#1}}}
\newcommand*{\sref}[1]{\S\ref{#1}}

\newcommand{\naive}{na\"ive }
\newcommand{\Naive}{Na\"ive }

\begin{document}

\begin{frontmatter}

\title{
BEAC: Imitating Complex Exploration and Task-oriented Behaviors for Invisible Object Nonprehensile Manipulation
}

\author[naistaddress,kobeaddress]{Hirotaka Tahara\corref{cor1}}\ead{ h-tahara@kobe-kosen.ac.jp}
\author[naistaddress]{Takamitsu Matsubara}\ead{takam-m@is.naist.jp}

\cortext[cor1]{Corresponding author}

\address[naistaddress]{Nara Institute of Science and Technology, 630-0192, Nara, Japan}
\address[kobeaddress]{Kobe City College of Technology, 651-2194, Hyogo, Japan}

\begin{abstract}
Applying imitation learning (IL) is challenging to nonprehensile manipulation tasks of invisible objects with partial observations, such as excavating buried rocks. The demonstrator must make such complex action decisions as exploring to find the object and task-oriented actions to complete the task while estimating its hidden state, perhaps causing inconsistent action demonstration and high cognitive load problems. For these problems, work in human cognitive science suggests that promoting the use of pre-designed, simple exploration rules for the demonstrator may alleviate the problems of action inconsistency and high cognitive load. Therefore, when performing imitation learning from demonstrations using such exploration rules, it is important to accurately imitate not only the demonstrator's task-oriented behavior but also his/her mode-switching behavior (exploratory or task-oriented behavior) under partial observation. Based on the above considerations, this paper proposes a novel imitation learning framework called Belief Exploration-Action Cloning (BEAC), which has a switching policy structure between a pre-designed exploration policy and a task-oriented action policy trained on the estimated belief states based on past history. In simulation and real robot experiments, we confirmed that our proposed method achieved the best task performance, higher mode and action prediction accuracies, while reducing the cognitive load in the demonstration indicated by a user study.
\end{abstract}


\end{frontmatter}

% \linenumbers
%%%%%%%%%%%%%%%%%%%%%%%%%%%%%%%%%%%%%%%%%%%%%%%%%%%%%%%%%%%%%%%%%%%%%%%%%%%%

%%%%%%%%%%%%%%%%%%%%%%%%%%%%%%%%%%%%%%%%%%%%%%%%%%%%%%%%%%%%%%%%%%%%%%%%%%%%
\section{Introduction}

\begin{figure}[tb]
\centering
\includegraphics[width=1.0\hsize] {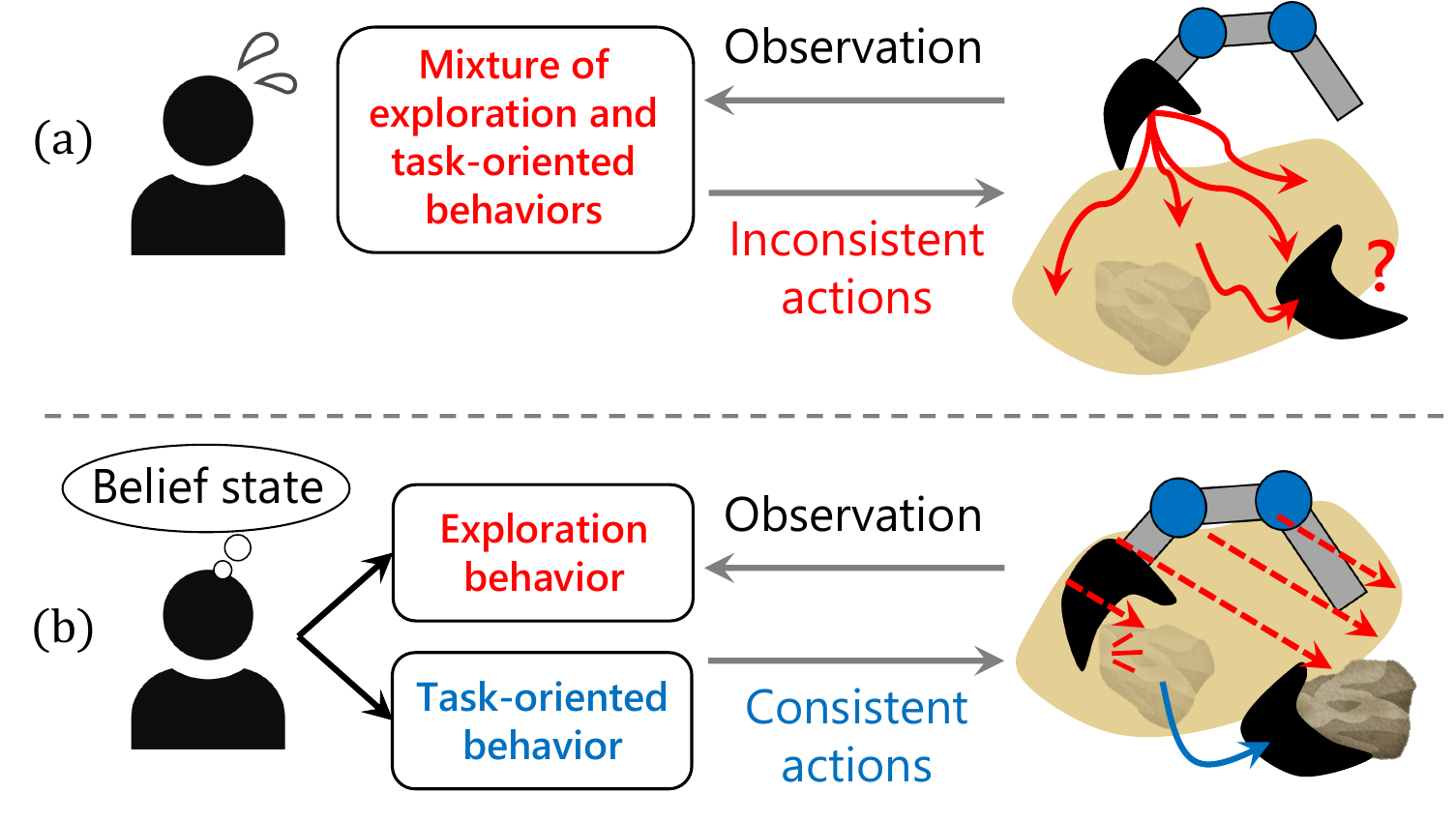}
\caption{
Invisible rock-removal task where rock is buried in sand (brown area): (a) In conventional imitation learning approaches, demonstrator has complex behavior that implicitly includes exploration and task-oriented behavior (red arrows) under partial observation. This results in inconsistent action demonstration and a high cognitive load, and the learned policy (blue arrow) fails. (b) Our approach explicitly introduces mode-switching behavior between exploration (red dotted arrows) and task-oriented behavior (blue arrow) in both demonstrator and learned policies. This makes action demonstration consistent while reducing cognitive load, and the learned policy succeeds.
}
\label{fig:introduction_problem}
\end{figure}

Imitation learning (IL) is a promising approach for automating robot tasks by learning policies that imitate human decision-making from demonstration data \cite{osa2018algorithmic}. Recent works have verified IL’s effectiveness for object manipulation tasks with full observability, where a target object is visible to the demonstrator \cite{zhang2018deep}. However, if the target object is invisible, such as for a buried rock excavation task, demonstrators must make such complex nonprehensile-action decisions as exploring to find the object and task-oriented actions to complete the task while estimating the invisible object’s hidden state under partial observation. This demonstrator's behavior may result in inconsistent action demonstration (\figref{fig:introduction_problem}(a)) \cite{subramanian2016exploration} and high cognitive load problems, leading to the deterioration of the learned policy's performance \cite{oh2021bayesian}.

Addressing the problem of cognitive load, a human cognitive science study \cite{alvarado2014reliance} argued that using non-optimal but heuristic exploration rules rather than complex behavior can reduce the cognitive load for situations with a lack of information or limited mental capabilities. Inspired by this conclusion, we can instruct demonstrators to utilize a pre-designed, simple exploration policy in imitation learning as a switching scheme of exploration and task-oriented action policies that reduce the demonstrator's cognitive load. Using a pre-designed exploration policy will also reduce the action inconsistency since automatic exploration can limit the demonstrator's arbitrary action decisions. Perhaps we can simultaneously deal with the above issues. For imitation learning in such a setting, accurately imitating both the demonstrator's task-oriented behavior and his/her mode-switching behavior (exploration or task-oriented behavior) is critical under partial observation. For this aspect, we hypothesize that instead of mode-switching and task-oriented action demonstrations based on instantaneous observations that lack sufficient information, the accuracy of mode and action predictions can be improved based on beliefs inferred from past history.

Based on the above considerations, we propose a novel imitation learning framework called \textit{Belief Exploration-Action Cloning} (\textbf{BEAC}), which imitates complex behaviors that include exploration and task-oriented actions for nonprehensile manipulation tasks of invisible objects (\figref{fig:introduction_problem}(b)). Specifically, BEAC has a mode-switching structure between a pre-designed exploration policy to acquire environmental information and a task-oriented action policy to achieve a task in both the demonstrator and the learned policy. Both the mode-switching and task-oriented action policies are trained with demonstrated mode and action labels using estimated belief states from a series of past observations and actions. Motivated by the states discussed in system identification \cite{katayama2005subspace, yamada2023subspace}, we maximize the mutual information between the current belief state and a multi-step future/past state for training a belief state estimator by following \cite{gangwani2020learning} to lead to more information in the belief states for improving the mode and action prediction accuracy.

We conducted evaluations in the nonprehensile manipulation tasks of invisible objects such as object pushing simulation task and a real robot buried-rock-removal task with human subjects. We confirmed that the proposed method achieved higher mode and action prediction accuracy and the best performance owing to a belief-based mode-switching policy structure, while reducing the cognitive load in demonstrations indicated by user surveys \cite{hart2006nasa}.
%%%%%%%%%%%%%%%%%%%%%%%%%%%%%%%%%%%%%%%%%%%%%%%%%%%%%%%%%%%%%%%%%%%%%%%%%%%%

%%%%%%%%%%%%%%%%%%%%%%%%%%%%%%%%%%%%%%%%%%%%%%%%%%%%%%%%%%%%%%%%%%%%%%%%%%%%
\section{Related works}

\subsection{Imitation learning under partial observation}

Imitation learning (IL) is a promising approach to obtain policies from demonstration data for automating robot tasks \cite{osa2018algorithmic}. Recent works have verified its effectiveness for several robot tasks that assume full observability in which states are available that contain sufficient information to determine appropriate actions \cite{zhang2018deep}. However, such real robot applications as rock removal earthwork tasks often suffer from partial observability \cite{lauri2022partially}. Gangwani et al. \cite{gangwani2020learning} pointed out that insufficient focus has been applied to IL in partially observable environments. They suggest that belief states must be estimated for learning optimal policies based on multi-step future and past information since \naive representation of belief states, which only consider past information, tend to ignore the relevant historical context. 

Their work represents partial observations by removing some observations from the state collected by a learned reinforcement learning policy, which behaves optimally like a demonstrator in the data collection phase. This assumption is problematic for nonprehensile manipulation skill of invisible objects because it ignores the difficulty of human demonstration, which is our focus. Since this issue makes the demonstration inconsistent \cite{subramanian2016exploration} and complicates policy learning \cite{oh2021bayesian}, our contribution is to focus on the issue of human demonstration for nonprehensile manipulation tasks of invisible objects. Please note that, as our aim is to explore imitation learning methods, reinforcement learning approaches that use belief states under partial observations are not within the scope of our research \cite{igl2018deep, zintgraf2020varibad}.

\subsection{Invisible object nonprehensile manipulation}

Invisible object manipulation tasks, including blind prehensile manipulation \cite{dang2011blind,felip2012contact,shaw2019robust,yang2020deep,murali2020learning,zhong2022soft} or blind non-prehensile manipulation \cite{zhou2023learning}, require exploration to find invisible objects based on force/tactile sensory data. Blind non-prehensile manipulation especially requires accurate state estimation of invisible objects to plan a non-grasping motion that examines the interaction of a robot, an object, and an environment. Since such tasks have complex dynamics that complicate applying analytical approaches, data-driven approaches have been studied in recent years \cite{zhou2023learning}. 

Zhou et al. \cite{zhou2023learning} employed a reinforcement learning approach for their non-prehensile object-pushing task. However, applying such an approach to our rock-removal task is difficult because trial-and-error exploration is cost expensive in real robot scenarios, and simulations are also unavailable due to the computational costs of contact among a huge amount of sand particles. Against these limitations, our study explores the imitation learning approach for blind non-prehensile manipulation tasks, which enables us to learn policies from a limited number of demonstration data collected by human demonstrators (Please note that studies on blind prehensile manipulation are not within the scope of our research).
%%%%%%%%%%%%%%%%%%%%%%%%%%%%%%%%%%%%%%%%%%%%%%%%%%%%%%%%%%%%%%%%%%%%%%%%%%%%

%%%%%%%%%%%%%%%%%%%%%%%%%%%%%%%%%%%%%%%%%%%%%%%%%%%%%%%%%%%%%%%%%%%%%%%%%%%%
\section{Preliminaries}

We formulated a conventional imitation learning approach that learns policies with \naive belief state estimation based on past history in a partially observable environment.

\subsection{Partially observable Markov decision process}

We assume the following partially observable Markov decision process (POMDP) \cite{lauri2022partially}: 
\begin{align}
    \mathcal{M}=\langle \mathcal{S}, \mathcal{A}, \mathcal{O}, \mathcal{T}, \mathcal{U} \rangle, 
\end{align}
where $\mathcal{S}$, $\mathcal{A}$, $\mathcal{O}$, $\mathcal{T}$, and $\mathcal{U}$ are the state space, the action space, the observation space, the environment's dynamics, and the observation function, respectively. State $\mathbf{s}_t \in \mathcal{S}$ at time step $t$ is a latent variable that is not directly observed. Give action $\mathbf{a}_t \in \mathcal{A}$, the next state is obtained as $\mathbf{s}_{t+1} \sim \mathcal{T}(\mathbf{s}_{t+1}|\mathbf{s}_t, \mathbf{a}_t)$, and an observation is obtained as $\mathbf{o}_{t+1} \sim \mathcal{U}(\mathbf{o}_{t+1}|\mathbf{s}_{t+1})$.

\subsection{\Naive belief state estimation considering past history}

\Naive belief state estimation approximates a latent state based on past history \cite{chung2015recurrent}. Filtering distribution $p(\mathbf{s}_t | \mathbf{b}_t)$ approximates current state $\mathbf{s}_t$ with belief state $\mathbf{b}_t$ as $p(\mathbf{s}_t | \mathbf{o}_{\leq t}, \mathbf{a}_{<t}) \approx p(\mathbf{s}_t | \mathbf{b}_t)$ based on the observations obtained up to current step $\mathbf{o}_{\leq t} = (\mathbf{o}_1, ... , \mathbf{o}_t)$ and the action taken up to previous step $\mathbf{a}_{< t} = (\mathbf{a}_0, ... , \mathbf{a}_{t-1})$. In this case, a belief state is obtained as $\mathbf{b}_{t+1} \sim \mathcal{T}(\mathbf{b}_{t+1}|\mathbf{b}_t, \mathbf{a}_t)$, and an observation is obtained as $\mathbf{o}_{t+1} \sim \mathcal{U}(\mathbf{o}_{t+1}|\mathbf{b}_{t+1})$. \Naive belief state $\mathbf{b}_t$ is obtained from belief state estimator $B_\phi(\mathbf{b}_t | \mathbf{o}_{\leq t}, \mathbf{a}_{<t})$ parameterized by $\phi$.

\subsection{Imitation learning (IL) using \naive belief states}
\label{conventional_IL}

The objective of a conventional IL under POMDP learns policy $\pi_{\theta^L}(\mathbf{a}_t| \mathbf{b}_t)$ that decides action $\mathbf{a}_t$ based on \naive belief state ${\mathbf{b}_t}$ that replicates a demonstrator's policy $\pi_{\theta^D}(\mathbf{a}_t| \mathbf{b}_t)$, which decides action $\mathbf{a}_t$ based on internal belief states $\mathbf{b}_t$ \cite{nguyen2016imitation}. Such a model is represented by a two-stage architecture: 1) a \naive belief state estimator and 2) a policy. Symbols $D$ and $L$ denote parameters related to the demonstrator and learned policies, and please note that the demonstrator is also denoted with the parameter to make the notation of the demonstrator and learned policies consistent for later formulation. 

To train the models, the demonstrator collects trajectories that consist of observation and action pairs: 
\begin{align}
\boldsymbol{\tau} = \left(\mathbf{o}_{1}, \mathbf{a}_{1}, \mathbf{o}_{2}, \mathbf{a}_{2}, \ldots, \mathbf{a}_{T-1}, \mathbf{o}_T\right),
\label{eq:trajectory}
\end{align}
where $T$ denotes the number of steps in a trajectory. Trajectory distribution $p(\boldsymbol{\tau})$ associated with demonstrator's policy $\pi_{\theta^D}(\mathbf{a}_t | \mathbf{b}_t)$ and demonstrator's belief state estimator $B_{\phi^D}(\mathbf{b}_t | \mathbf{o}_{\leq t}, \mathbf{a}_{<t})$ is defined as
\begin{align}
p(\boldsymbol{\tau}|\theta^D, \phi^D) = p(\mathbf{o}_1) \prod_{t=1}^{T} & p(\mathbf{o}_{t+1} | \mathbf{b}_{t+1}) p(\mathbf{b}_{t+1} | \mathbf{b}_{t}, \mathbf{a}_t) \pi_{\theta^D}(\mathbf{a}_t | \mathbf{b}_t) B_{\phi^D}(\mathbf{b}_t | \mathbf{o}_{\leq t}, \mathbf{a}_{<t}).
\label{eq:trajectory_distribution}
\end{align}
To replicate the demonstrator's policy and belief state estimator, the error of query policy $\pi_{\theta}$ and demonstrator's policy $\pi_{\theta^D}$ using trajectory $\boldsymbol{\tau}$ is defined as
\begin{align}
\mathcal{L}(\theta, \theta^D, \phi, \phi^D, \boldsymbol{\tau}) = \mathbb{E}_{p(\boldsymbol{\tau}|\theta^D, \phi^D)} \left\|\mathbf{a}_t - \pi_\theta(B_\phi(\mathbf{o}_{\leq t}, \mathbf{a}_{<t}))\right\|^2_2.
\label{eq:policy_loss}
\end{align}  
To minimize the expected surrogate loss as the objective function, parameters $\theta^L$ and  $\phi^L$ are acquired by solving the following optimization:
\begin{align}
\theta^L, \phi^L = \argmin_{\theta, \phi} \mathcal{L}(\theta, \theta^*, \phi, \phi^*, \boldsymbol{\tau}).
\label{eq:parameter_optimization}
\end{align}
%%%%%%%%%%%%%%%%%%%%%%%%%%%%%%%%%%%%%%%%%%%%%%%%%%%%%%%%%%%%%%%%%%%%%%%%%%%%

%%%%%%%%%%%%%%%%%%%%%%%%%%%%%%%%%%%%%%%%%%%%%%%%%%%%%%%%%%%%%%%%%%%%%%%%%%%%
\section{Proposed method}

\begin{algorithm}[htbp]
\footnotesize
\SetAlgoLined
\DontPrintSemicolon
\SetKwInOut{Parameter}{Parameter}
\SetKwInOut{Initialize}{Initialize}
\SetKwInOut{Inputs}{Inputs}

\Parameter{Demonstrator's mode-switching policy $\pi_{\lambda^D}$, task-oriented action policy $\pi^{\textrm{task-oriented}}_{\theta^D}$, belief state estimator $B_{\phi^D}$, learned mode-switching policy $\pi_{\lambda^L}$, task-oriented action policy $ \pi^{\textrm{task-oriented}}_{\theta^L}$, belief state estimator $B_{\phi^L}$. Exploration policy $\pi^{\textrm{exploration}}$ for both the demonstrator and learned policy.}

\Initialize{Set the total number of episodes $E$ and steps $T$.}

// Demonstration phase

\For{$e = 1$ to $E$}{
    \For{$t = 1$ to $T$}{
        Sample observation $\mathbf{o}_t$
        
        Belief state estimation: $\mathbf{b}_t = B_{\phi^D}(\mathbf{o}_{\leq t}, \mathbf{a}_{<t})$ 

        Mode-switching operation: $c_t = \pi_{\lambda^D}(\mathbf{b}_t)$

        \If{$c_t = 0$}{
            Exploration policy: $\mathbf{a}_t = \pi^{\textrm{exploration}}(\cdot)$
        } 
        \If{$c_t = 1$}{
            Task-oriented action policy: $\mathbf{a}_t = \pi^{\textrm{task-oriented}}_{\theta^D}(\mathbf{b}_t)$
        } 
    }
    Collected trajectories: $\boldsymbol{\tau}_{e} = \{ \mathbf{o}_{1:T}, \mathbf{a}_{1:T}, c_{1:T} \}$
}
Parameters $\lambda^L$, $\theta^L$, and $\phi^L$ are updated by \equref{eq:parameter_optimization_total} \\

// Testing phase

\For{$t = 1$ to $T$}{
    Sample observation $\mathbf{o}_t$
        
    Belief state estimation: $\mathbf{b}_t = B_{\phi^L}(\mathbf{o}_{\leq t}, \mathbf{a}_{<t})$ 

    Mode-switching operation: $c_t = \pi_{\lambda^L}(\mathbf{b}_t)$

    \If{$c_t = 0$}{
        Exploration policy: $\mathbf{a}_t = \pi^{\textrm{exploration}}(\cdot)$
    } 
    \If{$c_t = 1$}{
        Task-oriented action policy: $\mathbf{a}_t = \pi^{\textrm{task-oriented}}_{\theta^L}(\mathbf{b}_t)$
    } 
}

\caption{\footnotesize Belief Exploration-Action Cloning: BEAC}
\label{algorithm_beac}
\end{algorithm}

This section describes our proposed imitation learning framework called \textit{Belief Exploration-Action Cloning} (\textbf{BEAC}). It learns a mode-switching policy that switches between a pre-designed exploration policy and a task-oriented action policy using a belief state estimator trained with future and past information to improve the mode and action prediction accuracies.

First, we formulate an imitation model of complex behaviors that include exploration and a task-oriented action. Second, we formulate a belief state estimation based on future and past regularization. \algorithmref{algorithm_beac} outlines BEAC.

%------------------------------------------------------------
\subsection{Belief-based imitation model of complex behaviors that include exploration and task-oriented action}

\begin{figure}[tb]
\centering
\includegraphics[width=1.0\hsize]{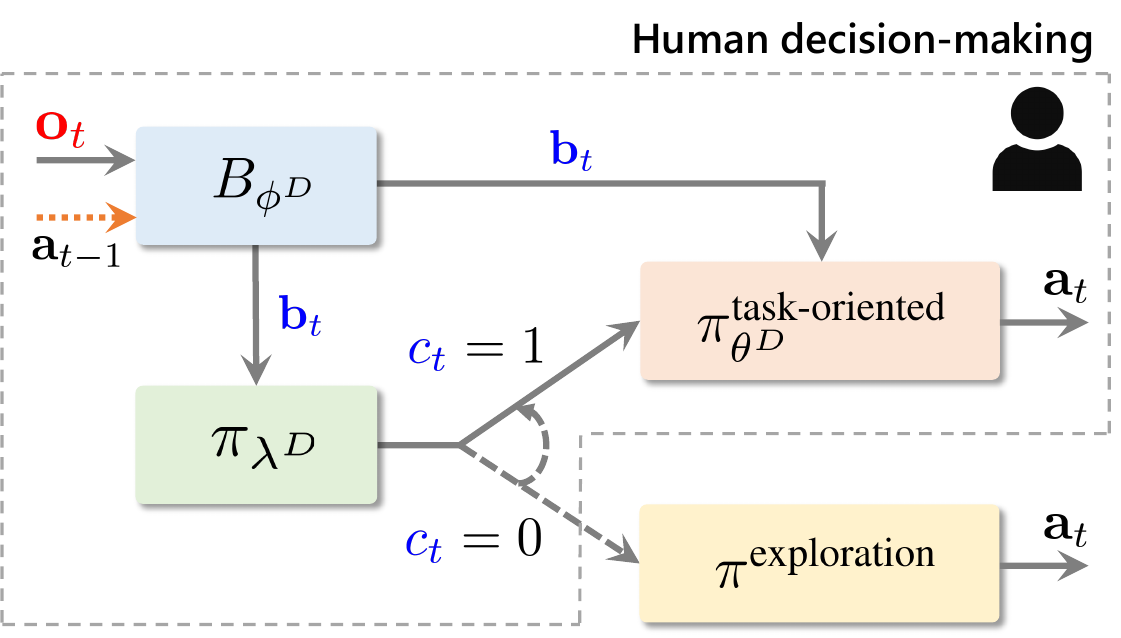}
\caption{
Block diagram of the proposed model on the demonstrator. The demonstrator's internal belief state estimator parameterized by $\phi^D$ estimates belief states based on the history of observations and actions. The demonstrator's mode-switching policy parameterized by $\lambda^D$ selects a non-parameterized exploration policy or task-oriented action policy parameterized by $\theta^D$ based on estimated belief state $\mathbf{b}_t$, where scalar value $c_t$ represents the mode label, which corresponds to either exploration or a task-oriented action.
}
\label{fig:proposed_method}
\end{figure}

\figref{fig:proposed_method} shows the demonstrator's model. The demonstrator's mode-switching policy parameterized by $\lambda^D$, which the selects exploration or task-oriented action policy based on internal belief state $\mathbf{b}_t$, is defined as $\pi_{\lambda^D}(c_t | \mathbf{b}_t)$, where scalar value $c_t$ represents the mode label, which corresponds to either exploration or a task-oriented action. The demonstrator's action policy, which chooses between exploration and a task-oriented action policy based on the mode, is defined as
\begin{align}
\pi_{\theta^D}(\mathbf{a}_t | \mathbf{b}_t, c_t) = 
\begin{cases}
\pi^{\textrm{task-oriented}}_{\theta^D}(\mathbf{a}_t | \mathbf{b}_t) & \text{if } c_t = 1\\
\pi^{\textrm{exploration}}(\mathbf{a}_t | \cdot) & \text{if } c_t = 0 \\
\end{cases}, 
\label{eq:semi-automatic_belief_policy}
\end{align}
where if mode $c_t = 0$, state-independent exploration policy $\pi^{\textrm{exploration}}$ is employed, and if $c_t = 1$, state-dependent task-oriented action policy $\pi^{\textrm{task-oriented}}$ is employed. While executing the exploration policy, the demonstrator pauses the manual task-oriented operation and focuses on monitoring the environment to decide the mode-switching timing. This approach reduces the cognitive load in demonstrations.

Unlike the conventional IL described in \sref{conventional_IL}, we assume that $c_t$ is collected through a demonstration (e.g., pressing a button to flip the mode) that resembles our previous work \cite{tahararal2023}. In contrast to the trajectory of \equref{eq:trajectory}, the trajectory in our framework is defined as observation, action, and mode pairs: 
\begin{align}
\boldsymbol{\tau}=\left(\mathbf{o}_{1}, \mathbf{a}_1, c_1, \ldots, \mathbf{a}_{T-1}, c_{T-1}, \mathbf{o}_T\right).
\label{eq:trajectory_with_mode}
\end{align}
Similar to the trajectory distribution of \equref{eq:trajectory_distribution}, our trajectory distribution which considers mode-switching is defined as
\begin{align}
p(\boldsymbol{\tau}|\theta^D, \lambda^D, \phi^D) = & p(\mathbf{o}_1) \prod_{t=1}^{T} p(\mathbf{o}_{t+1} | \mathbf{b}_{t+1}) p(\mathbf{b}_{t+1} | \mathbf{b}_{t}, \mathbf{a}_t) \nonumber \\ 
& \pi_{\theta^D}(\mathbf{a}_t | \mathbf{b}_t, c_t) \pi_{\lambda^D}(c_t | \mathbf{b}_t) B_{\phi^D}(\mathbf{b}_t | \mathbf{o}_{\leq t}, \mathbf{a}_{<t}).
\label{eq:trajectory_distribution_with_mode}
\end{align}

Following \equref{eq:semi-automatic_belief_policy}, the learned action policy is defined as
\begin{align}
\pi_{\theta^L}(\mathbf{a}_t | \mathbf{b}_t, c_t) = 
\begin{cases}
\pi^{\textrm{task-oriented}}_{\theta^L}(\mathbf{a}_t | \mathbf{b}_t) & \text{if } c_t = 1 \\
\pi^{\textrm{exploration}}(\mathbf{a}_t | \cdot) & \text{if } c_t = 0 \\
\end{cases}, \label{eq:learned_semi-automatic_belief_policy}
\end{align}
where belief state $\mathbf{b}_t$ is defined as $B_{\phi^L}(\mathbf{b}_t | \mathbf{o}_{\leq t}, \mathbf{a}_{<t})$, and the mode-switching policy is defined as $\pi_{\lambda^L}(c_t | \mathbf{b}_t)$. To replicate the demonstrator's action and mode-switching policies, we define the following objective functions using the query policy parameterized by $\theta$ and $\lambda$: 
\begin{align}
& \mathcal{L}_\textrm{action} (\theta, \theta^D, \lambda, \lambda^D, \phi, \phi^D, \boldsymbol{\tau}) = \nonumber \\
& \mathbb{E}_{p(\boldsymbol{\tau} | \theta^D, \lambda^D, \phi^D)} \Big[ c_t \big{\|} \mathbf{a}_t - \pi_{\theta}(B_\phi(\mathbf{o}_{\leq t}, \mathbf{a}_{<t}), \mathbf{c}_t) \big{\|}_2^2 \Big], 
\label{eq:action_loss} \\
& \mathcal{L}_\textrm{mode} (\theta, \theta^D, \lambda, \lambda^D, \phi, \phi^D, \boldsymbol{\tau}) = \nonumber \\  
& \mathbb{E}_{p(\boldsymbol{\tau} | \theta^D, \lambda^D, \phi^D)} \Big[ -c_t \log \pi_{\lambda} \big(B_\phi(\mathbf{o}_{\leq t}, \mathbf{a}_{<t})\big) - \left(1-c_t\right) \log \left(1-\pi_{\lambda}\big(B_\phi(\mathbf{o}_{\leq t}, \mathbf{a}_{<t})\big) \right) \Big]. 
\label{eq:mode-switching-loss}
\end{align}
\equref{eq:action_loss} and \equref{eq:mode-switching-loss} represent the action prediction loss and the mode prediction loss.
To learn a task-oriented action policy, \equref{eq:action_loss} has multiplier $c_t$ at the L2 loss to ignore the exploration data.

%------------------------------------------------------------
\subsection{Belief state estimation based on future and past regularizations}

\begin{figure}[tb]
\centering
\includegraphics[width=1.0\hsize] {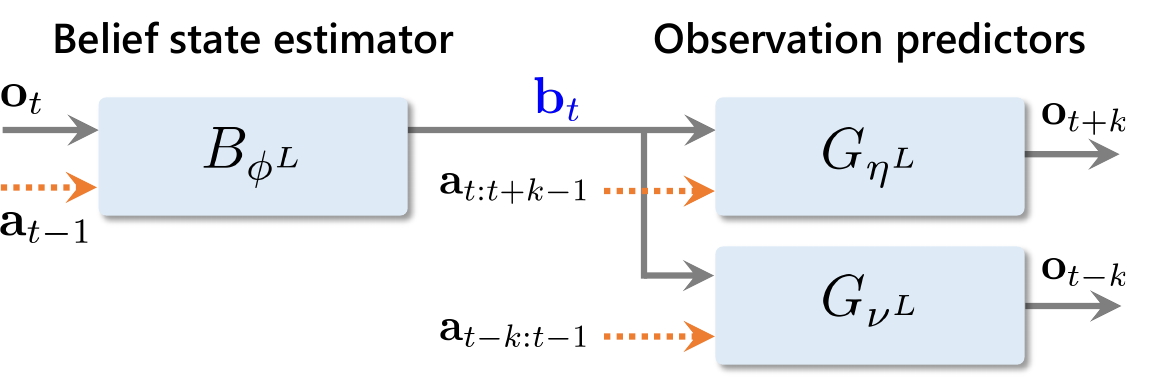}
\caption{
Belief state estimator with future and past regularization that considers k-step future and past information.  The autoencoder model has an encoder (recurrent natural network-based belief state estimator) and decoders (future and past observation predictors).
}
\label{fig:proposed_belief_regularization}
\end{figure}

To deal with nonprehensile manipulation tasks of invisible objects, \naive belief state representation, which only considers past information, tends to ignore the relevant historical context, and reacting to short-term past observations leads to incorrect mode and action predictions. Instead, we focus on the definition of the state in control theory, which compresses the future and past information described in system identification \cite{katayama2005subspace, yamada2023subspace} to obtain robust belief state representation. To achieve this state, we employ regularizations that consider future and past information in the training of a belief state estimator by following \cite{zintgraf2020varibad, gangwani2020learning}. We maximize the mutual information between current belief states and multi-step future and past states, which leads to informative belief states \cite{gangwani2020learning}. As shown in \figref{fig:proposed_belief_regularization}, the autoencoder model implements this regularization, where encoder $B_\phi$ is a recurrent neural network-based belief state estimator and two decoders are future and past observation predictors.

First, we formulate the future regularization that maximizes the mutual information between estimated belief state $\mathbf{b}_t$ and $k$-step future state. The correlation between future state $\mathbf{s}_{t+k}$ and current belief state $\mathbf{b}_t$ is obtained as conditional mutual information $I(\mathbf{b}_t ; \mathbf{s}_{t+k} | \mathbf{a}_{t:t+k-1})$ given action sequence $\mathbf{a}_{t:t+k-1}$. Since conditional independence $\mathbf{o}_{t+k} \indep \mathbf{b}_t|\mathbf{s}_{t+k}$ holds from the observation generation process under partial observation, the following inequality is obtained using data processing inequality:
\begin{align}
    I(\mathbf{b}_t ; \mathbf{s}_{t+k} | \mathbf{a}_{t:t+k-1}) \geq  I(\mathbf{b}_t ; \mathbf{o}_{t+k} | \mathbf{a}_{t:t+k-1}).
    \label{eq:inequality_mutual_information_future}
\end{align}
From the definition of mutual information, the lower bound of \equref{eq:inequality_mutual_information_future} is expanded:
\begin{align}
    &I(\mathbf{b}_t ; \mathbf{o}_{t+k} | \mathbf{a}_{t:t+k-1}) = \mathbb{E}_{\mathbf{a}_{t:t+k-1}} \Big[ H\big(\mathbf{o}_{t+k} | \mathbf{a}_{t:t+k-1}\big) - H\big(\mathbf{o}_{t+k} | \mathbf{b}_t, \mathbf{a}_{t:t+k-1}\big) \Big].
\end{align}
Since computing the above equation is difficult if $\mathbf{b}_t$ is high-dimensional, a variational approximation is applied based on mutual information maximization \cite{poole2019variational} to obtain the following variational lower bound: 
\begin{align}
    &I(\mathbf{b}_t ; \mathbf{o}_{t+k} | \mathbf{a}_{t:t+k-1}) \nonumber \\
    & \geq \mathbb{E}_{\mathbf{a}_{t:t+k-1}} \Big[ H(\mathbf{o}_{t+k}|\mathbf{a}_{t:t+k-1}) + \mathbb{E}_{\mathbf{o}_{t+k}, \mathbf{b}_t} \big[ \log q (\mathbf{o}_{t+k} | \mathbf{b}_t, \mathbf{a}_{t:t+k-1}) \big] \Big],
    \label{eq:lower_bound_mutual_information}
\end{align}
where $q$ denotes the variational distribution. We approximately compute left hand side of the above equation by maximizing the above variational lower bound as a surrogate function. Ignoring terms that do not include the parameter to be optimized, we maximize the following objective:
\begin{align}
    \max\limits_{{\phi, q}} &\mathbb{E}_{\mathbf{o}_{t+k}, \mathbf{b}_t, \mathbf{a}_{t:t+k-1}} \big[\log q(\mathbf{o}_{t+k} | B_{\phi^L}(\mathbf{b}_t | \mathbf{o}_{\leq t}, \mathbf{a}_{<t}), \mathbf{a}_{t:t+k-1})\big]
    \label{eq:maximization_mutual_information_future}.
\end{align}
Assuming a Gaussian distribution with a fixed variance and a mean expressed by the output of learned future decoder model $G_{\eta^L}$ parameterized by $\eta^L$ in variational distribution $q$, we obtain the following loss function:
\begin{align}
    \mathcal{L}_{\textrm{future}} = & \mathbb{E}_{p(\boldsymbol{\tau}|\theta^D, \lambda^D, \phi^D)} \big{\|} \mathbf{o}_{t+k} - G_{\eta^L}(B_{\phi^L}(\mathbf{o}_{\leq t}, \mathbf{a}_{<t}), \mathbf{a}_{t:t+k-1}) \big{\|} ^2_2
    \label{eq:loss_future_regularization}.
\end{align}
With the same procedure, the following is the loss function for maximizing the mutual information between the estimated belief state and the $k$-step past state:
\begin{align}
    \mathcal{L}_{\textrm{past}} = & \mathbb{E}_{p(\boldsymbol{\tau}|\theta^D, \lambda^D, \phi^D)} \big{\|} \mathbf{o}_{t-k} - G_{\nu^L}(B_{\phi^L}(\mathbf{o}_{\leq t}, \mathbf{a}_{<t}), \mathbf{a}_{t-k:t-1}) \big{\|}^2_2
    \label{eq:loss_past_regularization},
\end{align}
where $\nu^L$ is a parameter of past decoder model $G_{\nu^L}$.

%------------------------------------------------------------
\subsection{Parameter optimization}

To train all the parameters, the total loss is obtained by combining 
Eq.~\eqref{eq:action_loss}, \eqref{eq:mode-switching-loss}, \eqref{eq:loss_future_regularization}, and \eqref{eq:loss_past_regularization}:
\begin{align}
    \mathcal{L}_{\textrm{total}} = \mathcal{L}_{\textrm{action}} + \alpha \mathcal{L}_{\textrm{mode}} + \beta \mathcal{L}_{\textrm{future}} + \gamma \mathcal{L}_{\textrm{past}},
    \label{eq:total_loss}
\end{align}
where $\alpha$, $\beta$, and $\gamma$ are the weight for each loss. Finally, the parameters are acquired by solving the equation below:
\begin{align}
\theta^L, \lambda^L, \phi^L, \eta^L, \nu^L = \argmin_{\theta, \lambda, \phi, \eta, \nu} \mathcal{L}_{\textrm{total}}.
\label{eq:parameter_optimization_total}
\end{align}
%%%%%%%%%%%%%%%%%%%%%%%%%%%%%%%%%%%%%%%%%%%%%%%%%%%%%%%%%%%%%%%%%%%%%%%%%%%%

%%%%%%%%%%%%%%%%%%%%%%%%%%%%%%%%%%%%%%%%%%%%%%%%%%%%%%%%%%%%%%%%%%%%%%%%%%%%
\section{Evaluation}

\subsection{Overview}

In this section, we conducted experiments using nonprehensile manipulation tasks of invisible objects, such as object pushing simulation task and a real robot buried-rock-removal task to verify the effectiveness of the proposed method. Since these tasks involve complex dynamics in the physical interactions among the robot's end-effector, an object, and the environment, a data-driven approach is a reasonable alternative to analytical schemes. The following two questions are clarified in this verification:
\begin{itemize}
\item[Q1] Does the proposed method improve the task performance, the action prediction accuracy, and the mode prediction accuracy owing to the belief-based imitation model and belief state estimation with future and past regularization? Since task performance is affected by both mode prediction accuracy and action prediction accuracy, we evaluate task performance and action prediction accuracy separately.
\item[Q2] Does it reduce the cognitive load in the demonstration owing to the switching policy structure?
\end{itemize}
Q1 is comprehensively verified in a simulation using an algorithmic demonstrator, and Q2 is verified in a real robot experiment involving actual human subjects.

\subsection{Comparison methods}

\begin{table}[tb]
\centering
\caption{
This table shows comparison methods and their characteristics. Ours is the proposed method. Ours w/o past, Ours w/o future, and Ours w/o reg are the ablations regarding belief state regularization, and BC w/ switch, BC w/ belief, and BC are baseline methods.
}
\label{table:evaluation_comparison}
\begin{tabular}{l | c c c c}
\hline
\diagbox[width=2.7cm, height=\line]{\ }{\ } & Mode switching & State estimation & Future reg. & Past reg. \\
\hline
\hline
Ours & $\checkmark$ & $\checkmark$ & $\checkmark$ & $\checkmark$ \\
Ours w/o past & $\checkmark$ & $\checkmark$ & $\checkmark$ & $--$ \\
Ours w/o future & $\checkmark$ & $\checkmark$ & $--$ & $\checkmark$ \\
Ours w/o reg & $\checkmark$ & $\checkmark$ & $--$ & $--$ \\
BC w/ switch & $\checkmark$ & $--$ & $--$ & $--$ \\
BC w/ belief & $--$ & $\checkmark$ & $--$ & $--$ \\
BC & $--$& $--$ & $--$ & $--$ \\
\hline
\end{tabular}
\end{table}

The comparison methods are shown in \tableref{table:evaluation_comparison}. Each is categorized based on four characteristics: 1) mode switching, 2) (belief) state estimation, 3) future regularization, and 4) past regularization. For the proposed method (Ours), three ablations (Ours w/o past, Ours w/o future, and Ours w/o reg) are categorized in terms of belief state regularizations with characteristics 3 and 4. As baseline methods, we also compared the Behavior Cloning (BC) methods with characteristic 1 alone (BC w/ switch), with characteristic 2 alone (BC w/ belief), and without these characteristics (BC).

%%%%%%%%%%%%%%%%%%%%%%%%%%%%%%%%%%%%%%%%%%%%%%%%%%%%%%%%%%%%%%%%%%%%%%%%%%%%

%%%%%%%%%%%%%%%%%%%%%%%%%%%%%%%%%%%%%%%%%%%%%%%%%%%%%%%%%%%%%%%%%%%%%%%%%%%%
\subsection{Simulation experiment}
\label{sec:simulation}

\subsubsection{Task setting}

\begin{figure}[htbp]
\centering
\includegraphics[width=0.9\hsize] {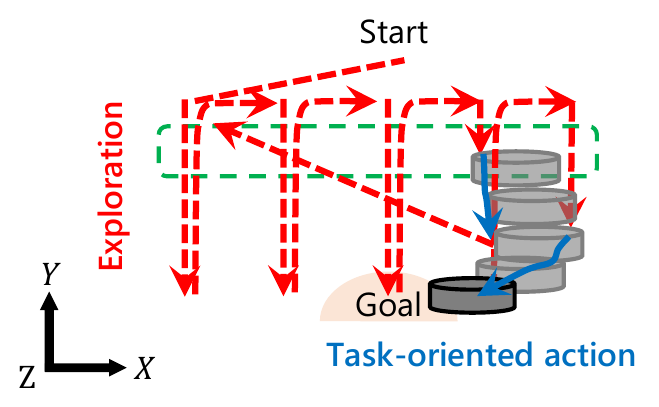}
\caption{
Simulated robotic environment for invisible object's nonprehensile pushing task. Red dotted arrows and blue arrow represent exploration (forward and backward direction) and task-oriented action.
}
\label{fig:evaluation_simulation_environment}
\end{figure}

\figref{fig:evaluation_simulation_environment} shows the simulated robotic environment for the invisible objects' nonprehensile pushing task. We built it on the customized OpenAI Gym's Fetch robotics environment. In this task, the robotic arm pushes a black cylinder object (on a table), which is assumed to be an invisible object, toward the goal (a red ball whose perimeter is $\pm 10~\textrm{cm}$). Uniform noise $\mathcal{U}(-\sigma, \sigma)$ with $\sigma = 10~\textrm{cm}$ (area surrounded by the green dotted line) is added to the object's initial position ($X$ and $Y$) as the task's randomness. The task succeeds when the distance between the object and the goal is equal to or less than the threshold value ($< 10~\textrm{cm}$).

\subsubsection{Demonstration setting}

Since this simulation experiment wants to confirm the task performance and the mode and action prediction accuracies, we used an algorithmic controller instead of a human demonstrator to provide comprehensive validation and avoid the degradation of demonstration quality. We assume that the algorithmic demonstrator can access the object's oracle state to ensure that the demonstration can be optimally conducted, as in a previous work \cite{gangwani2020learning}. For demonstration with switching, the demonstrator collects information about the invisible object using an empirically designed automatic exploration policy \cite{murali2020learning}. Based on the state (the haptic information and the distance between the object and the end-effector), the demonstrator switches to manual operation for the task-oriented actions. In contrast, we use the same motion without arbitrary exploration and mode label collection for BC and BC w/ belief since this is the upper bound of the performance of these policies. The observations used for policy learning are the position $(X, Y, Z)$ of the robot's end-effector, the end-effector's movement speed $(\dot{X},\dot{Y},\dot{Z})$, and the 6-D force torque sensor value which is attached to the robot's wrist. The predicted action is the end-effector's position deviation $(\Delta{X},\Delta{Y},\Delta{Z})$ at each step.

\subsubsection{Learning setting}

Five sets of models were trained with random seeds after collecting 100 demonstration trajectories. Each model was tested ten times. We used a four-layer deep neural network with an input layer, two 64-dimensional intermediate layers, and an output layer for the action policy. A three-layer deep neural network with an input layer, a 64-dimensional intermediate layer, and an output layer was used for the mode-switching policy. The encoder model used for the belief state estimation was a 64-dimensional LSTM layer (LSTM is used here for ease of implementation, although other time series models (e.g., Transformer) can be used.). The decoder model for the belief state regularization uses a three-layer deep neural network with an input layer, a 64-dimensional intermediate layer, and an output layer.

\subsubsection{Result}

\begin{table}[tb]
\centering
\caption{
This table shows the learned policy's success rate, mode prediction (pred.) accuracy (acc.), and action prediction (pred.) loss in each method. Please note that each value of action prediction loss has [$\times 10^{-5}$], which is omitted for space limitation.
}
\label{table:evaluation_simulation}
\begin{tabular}{l | c c c}
\hline
\diagbox[width=2.7cm, height=\line]{\ }{\ } & Success rate [\%] & Mode pred. acc. [\%] & Action pred. loss \\
\hline
\hline
Ours & $\mathbf{88 \pm 7.5}$ & $\mathbf{97.4 \pm 2.69}$ & $\mathbf{0.87 \pm 0.09}$ \\
Ours w/o past & $78 \pm 20.4$ & $95.5 \pm 1.41$ & $2.04 \pm 0.42$\\
Ours w/o future & $72 \pm 9.8$ & $95.0 \pm 0.28$ & $2.81 \pm 1.56$ \\
Ours w/o reg & $66 \pm 4.9$ & $94.6 \pm 1.84$ & $3.25 \pm 1.58$ \\
BC w/ switch & $44 \pm 4.9$ &  $82.8 \pm 0.85$ &  $9.01 \pm 1.35$ \\
BC w/ belief & $26 \pm 20.6$ & $--$ & $8.19 \pm 0.17$ \\
BC & $6 \pm 4.9$ & $--$ & $12.1 \pm 0.28$ \\
\hline
\end{tabular}
\end{table}

\begin{figure}[htbp]
    \centering
    
    \begin{minipage}[b]{0.24\linewidth}
    \centering
    \includegraphics[width=1.0\hsize]{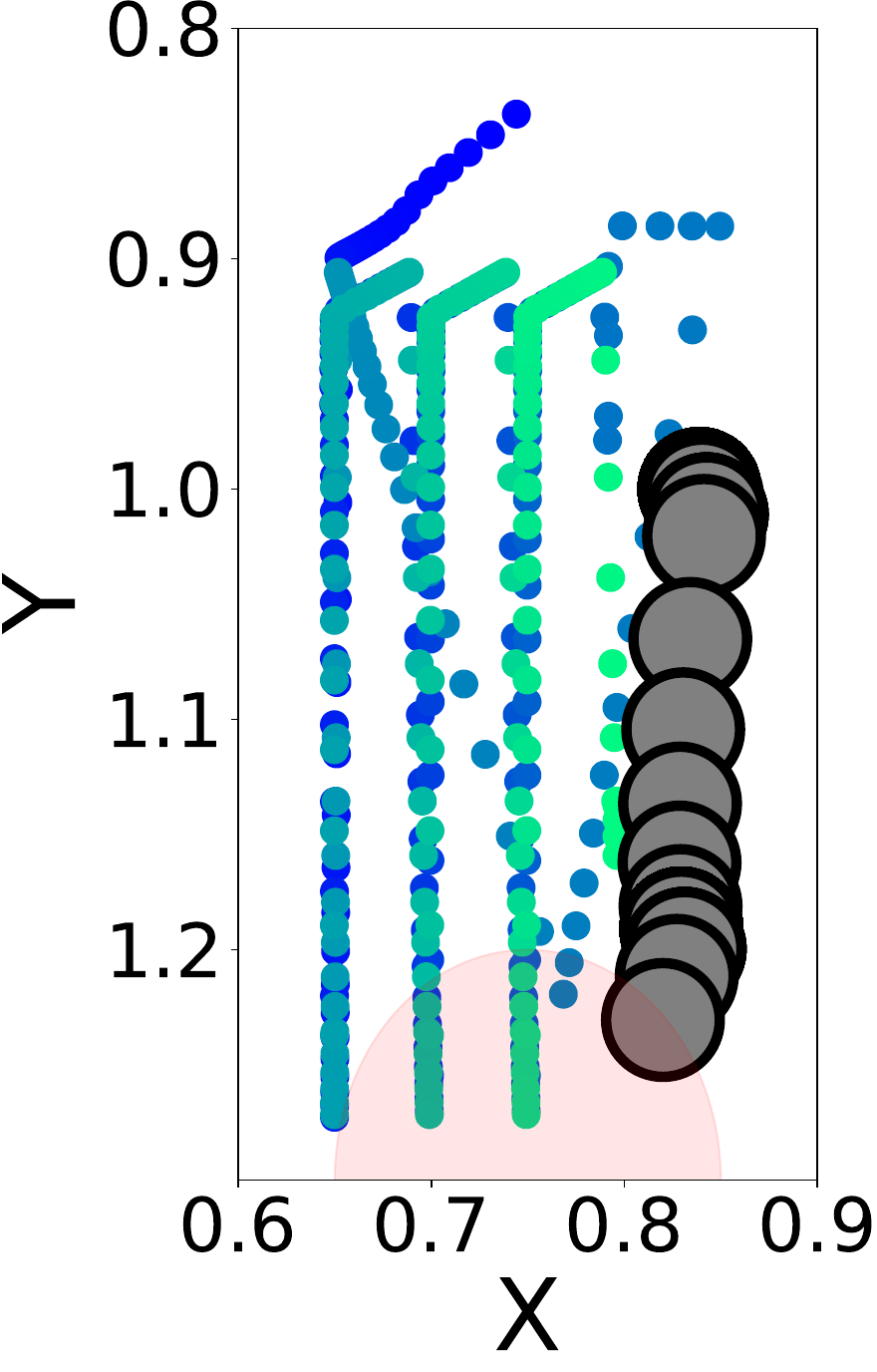}
    \subcaption{Demo}
    \end{minipage}    
    \begin{minipage}[b]{0.24\linewidth}
    \centering
    \includegraphics[width=1.0\hsize]{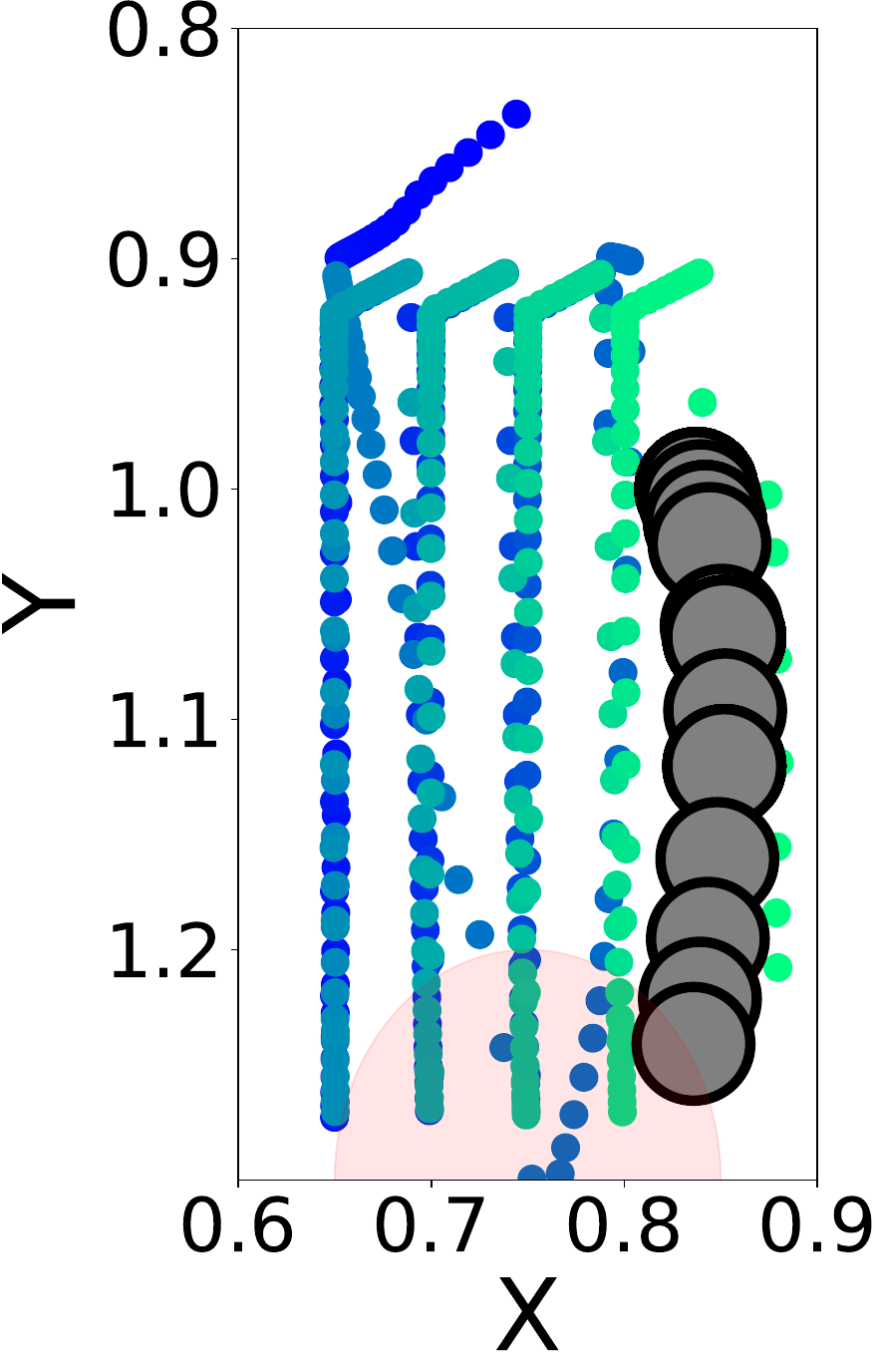}
    \subcaption{Ours}
    \end{minipage}

    \begin{minipage}[b]{0.24\linewidth}
    \centering
    \includegraphics[width=1.0\hsize]{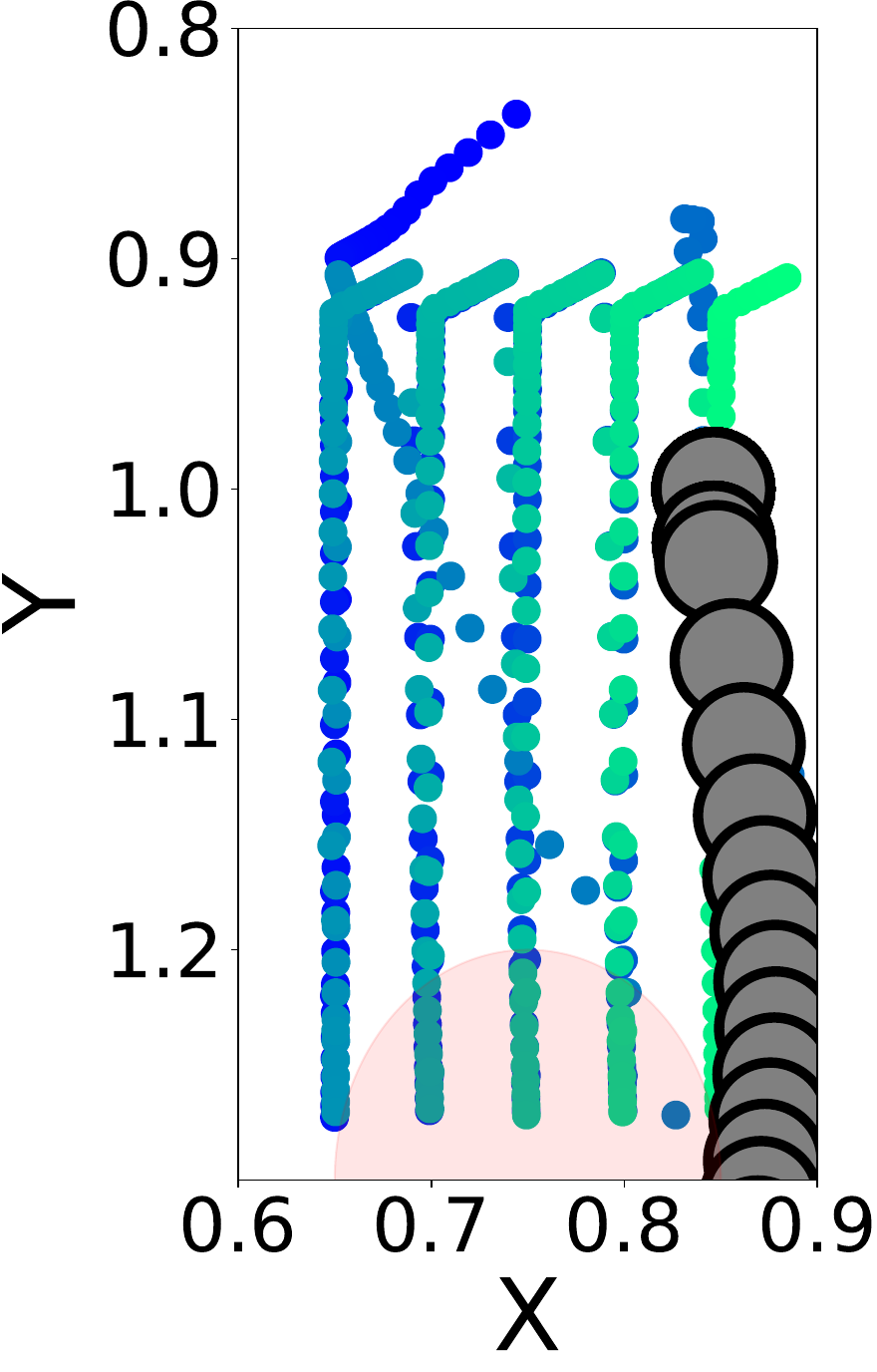}
    \subcaption{Ours w/o past}
    \end{minipage}
    \begin{minipage}[b]{0.24\linewidth}
    \centering
    \includegraphics[width=1.0\hsize]{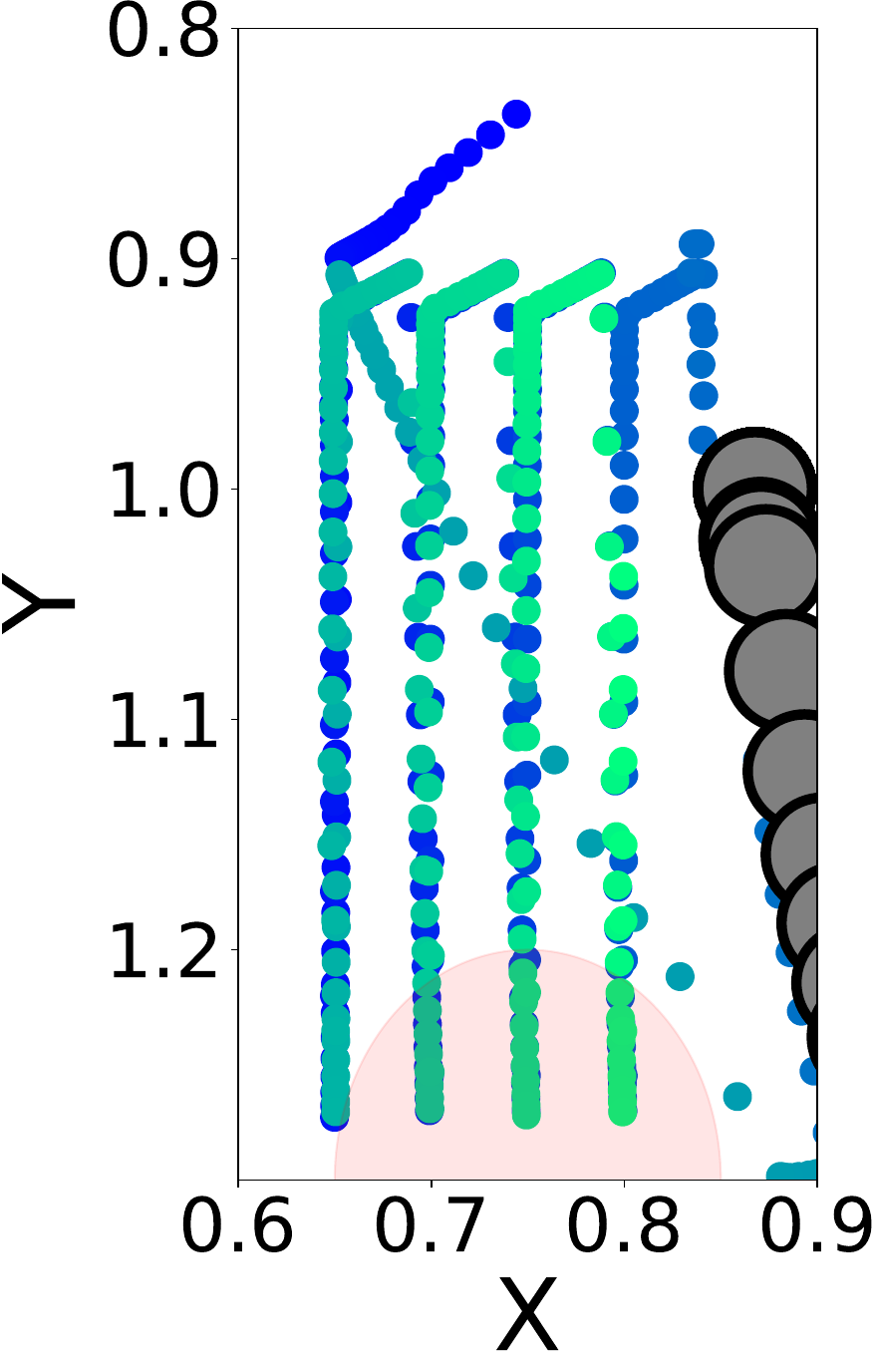}
    \subcaption{Ours w/o future}
    \end{minipage}
    \begin{minipage}[b]{0.24\linewidth}
    \centering
    \includegraphics[width=1.0\hsize]{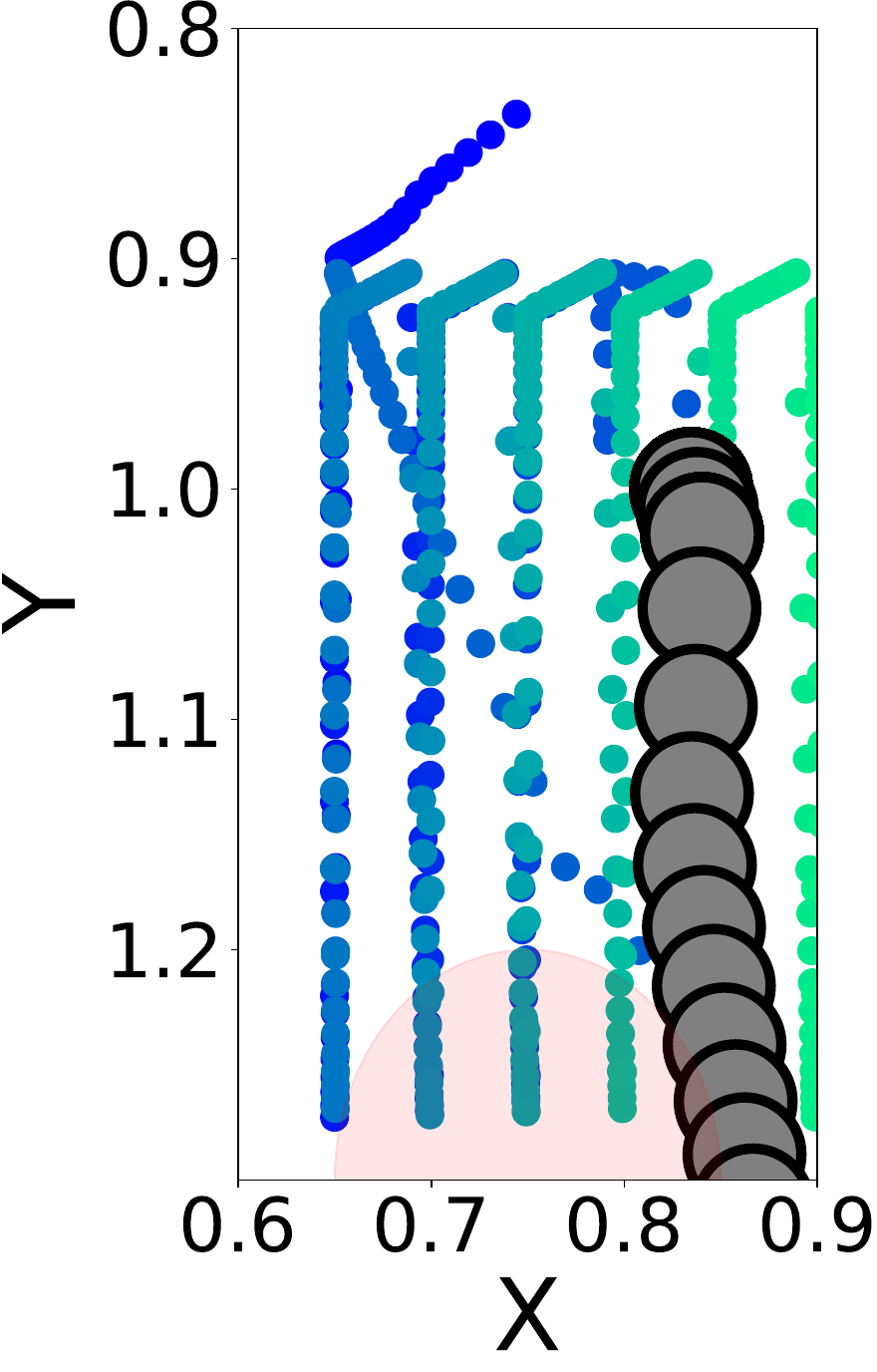}
    \subcaption{Ours w/o reg}
    \end{minipage}

    \begin{minipage}[b]{0.24\linewidth}
    \centering
    \includegraphics[width=1.0\hsize]{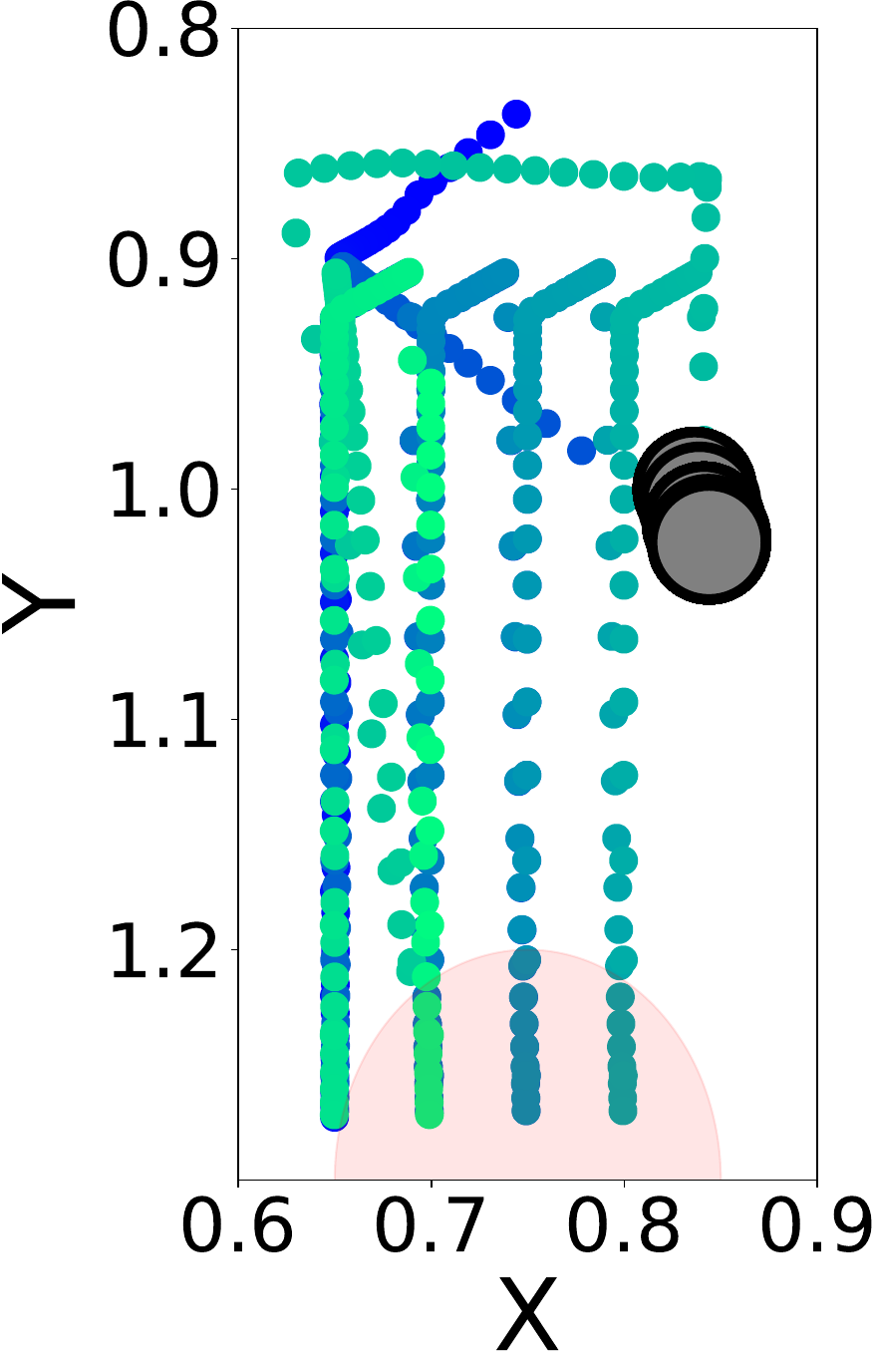}
    \subcaption{BC w/ switch}
    \end{minipage}
    \begin{minipage}[b]{0.24\linewidth}
    \centering
    \includegraphics[width=1.0\hsize]{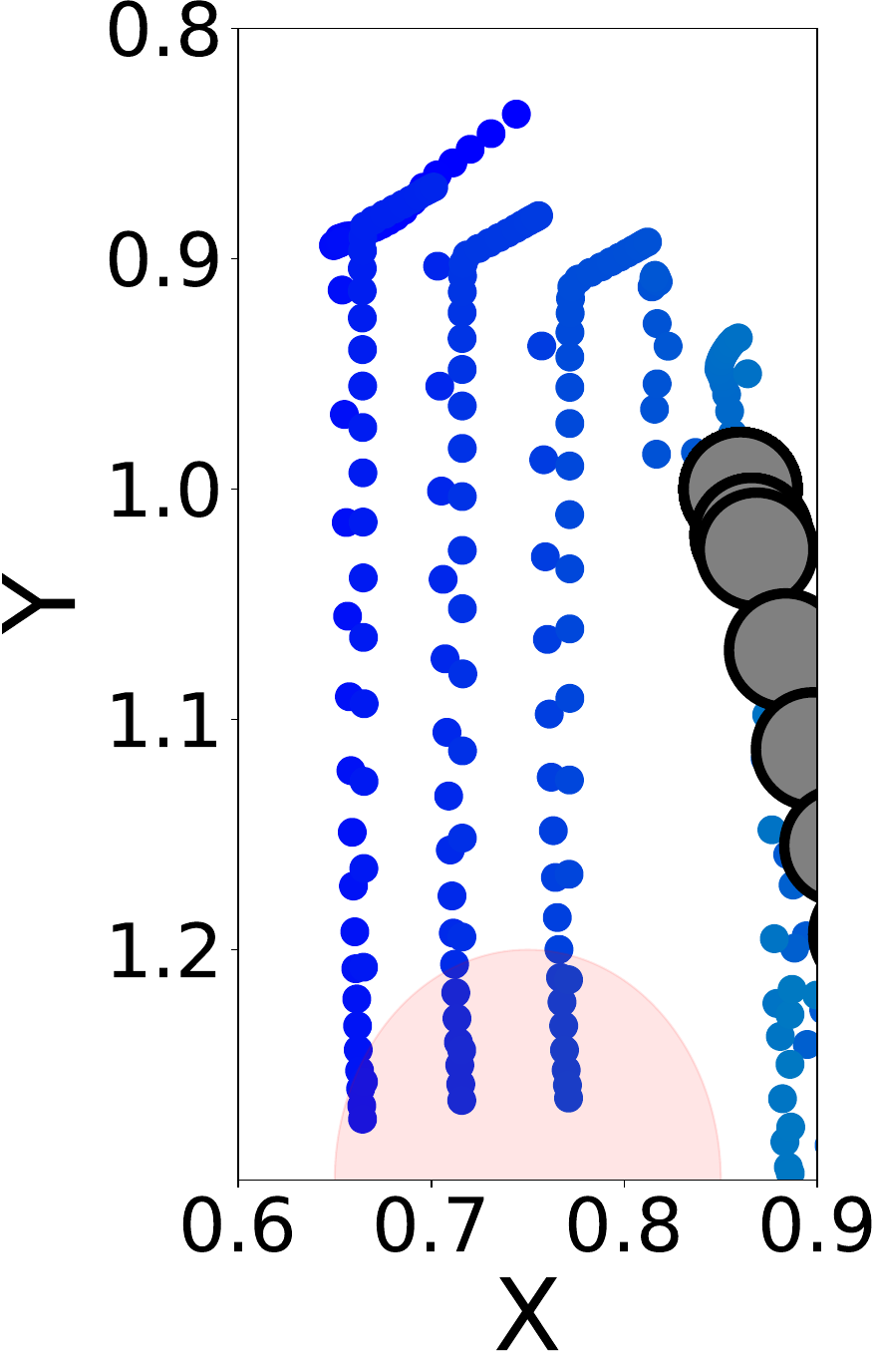}
    \subcaption{BC w/ belief}
    \end{minipage}
    \begin{minipage}[b]{0.24\linewidth}
    \centering
    \includegraphics[width=1.0\hsize]{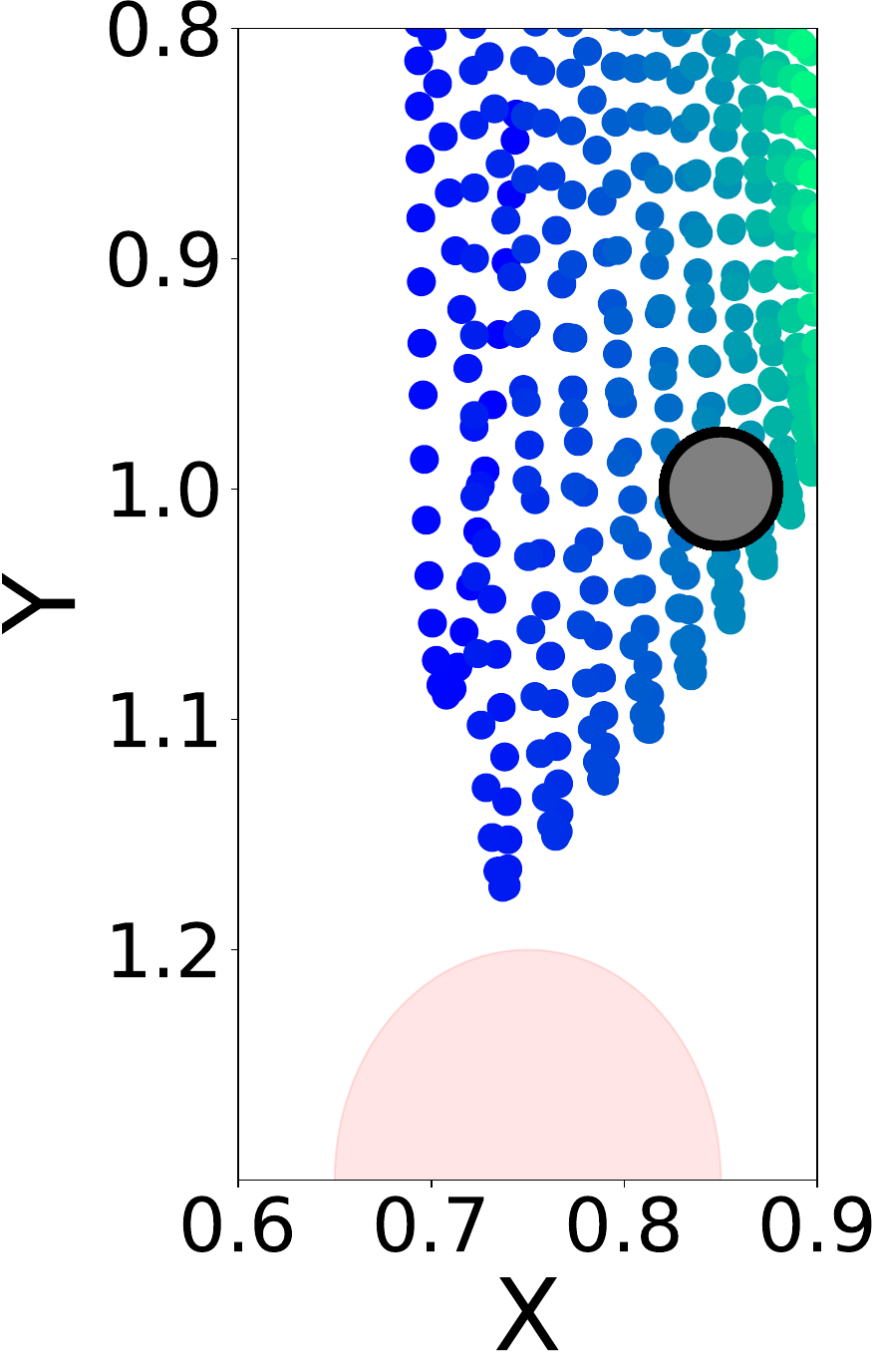}
    \subcaption{BC}
    \end{minipage}

\caption{
Results of test trajectories visualized in $X$-$Y$ 2D-plane when initial object position is fixed to a certain place: Black circle trajectory represents object's position. Color transition (dark blue to light green) on the end-effector trajectory represents a time-series transition. Pink shaded area shows goal area. Demonstration and Ours only achieved task with a similar object trajectory.
}
\label{fig:evaluation_simulation_trajectory}
\end{figure}

The success rate, the mode prediction accuracy, and the action prediction loss of the learned policies are shown in \tableref{table:evaluation_simulation}. Ours achieved the best success rate against the other comparison methods. Compared to the ablation methods regarding regularization (Ours w/o past, Ours w/o future, and Ours w/o reg), we confirmed that introducing future and past regularization in the belief state estimator improved the success rate. This result is also shown by Ours, which has the highest mode prediction accuracy and the lowest action prediction loss. In addition, Ours and the ablation methods indicated the effectiveness of utilizing automatic exploration policy and belief state estimation since they outperformed baseline methods BC w/ switch, BC w/ belief, and BC in terms of success rate and action prediction loss. These performances were reflected in the test trajectories shown in \figref{fig:evaluation_simulation_trajectory}. For comparison, we show them with the same initial object's position. Ours achieved a trajectory that resembled the demonstrated trajectory, unlike the other methods, which failed to push the object toward the goal.

\begin{table}[tb]
\centering
\caption{
Success rate of learned policies with $k=20, 10, 3$ considered in belief state regularizations in the simulation experiment:
Last row shows training times of models on a PC (CPU: AMD Ryzon 9, GPU: RTX3090).
}
\label{table:evaluation_simulation_performance_ablation}
\begin{tabular}{l | c | c | c}
\hline
\diagbox[width=2.5cm, height=\line]{\ }{\ } & \multicolumn{3}{c}{Success rate [\%]} \\
\hline
Method & $k=20$ & $k=10$ & $k=3$ \\
\hline
\hline
\textbf{Ours}   & $\mathbf{86 \pm 8.0}$     & $\mathbf{88 \pm 7.5}$     & $\mathbf{80 \pm 16.7}$\\
Ours w/o past   & $76 \pm 15.0$             & $78 \pm 20.4$             & $72 \pm 19.4$ \\
Ours w/o future & $70 \pm 12.6$             & $72 \pm 9.8$              & $64 \pm 22.4$ \\
\hline
\hline
Training time   & 4146 [sec]                 & 2877 [sec]                 & 1655 [sec] \\                   
\hline
\end{tabular}
\end{table}

\begin{figure}[htbp]
    \centering
    
    \begin{minipage}[b]{1.0\linewidth}
    \centering
    \includegraphics[width=1.0\hsize]{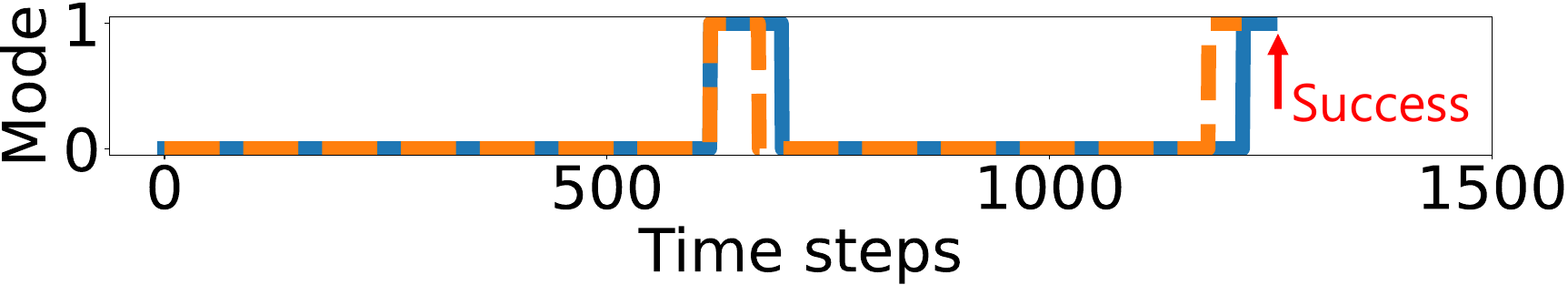}
    \subcaption{Ours}
    \end{minipage}
    
    \begin{minipage}[b]{1.0\linewidth}
    \centering
    \includegraphics[width=1.0\hsize]{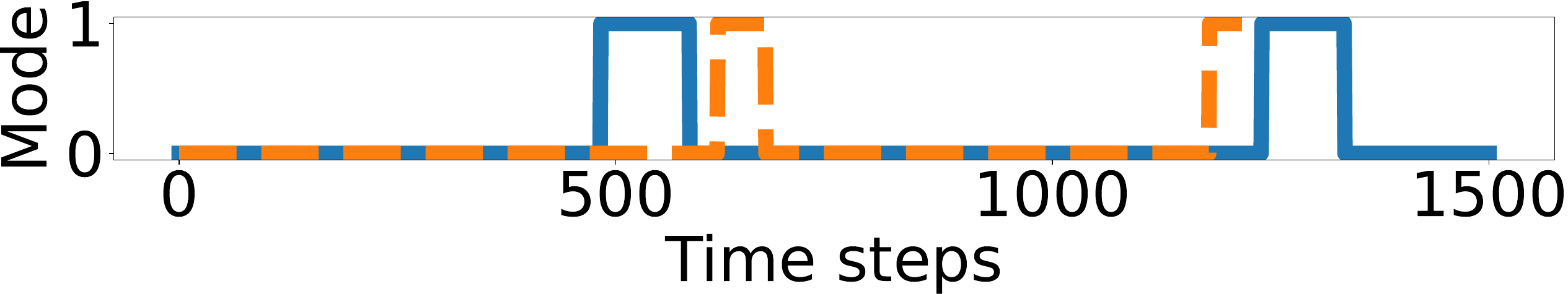}
    \subcaption{Ours w/o past}
    \end{minipage}
    
    \begin{minipage}[b]{1.0\linewidth}
    \centering
    \includegraphics[width=1.0\hsize]{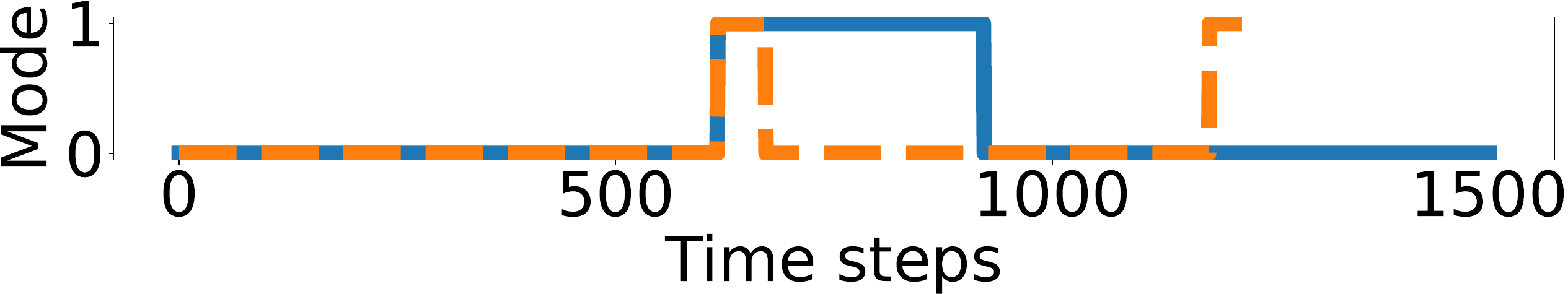}
    \subcaption{Ours w/o future}
    \end{minipage}
    
    \begin{minipage}[b]{1.0\linewidth}
    \centering
    \includegraphics[width=1.0\hsize]{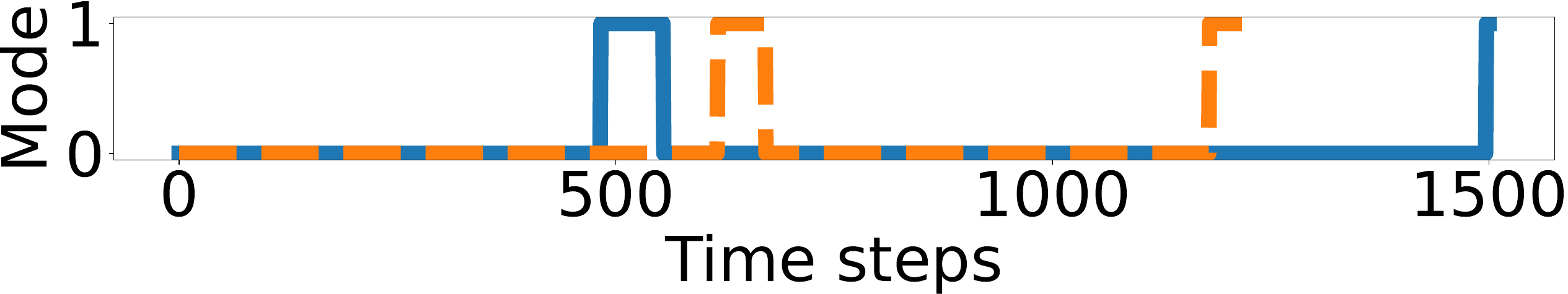}
    \subcaption{Ours w/o reg}
    \end{minipage}
    
    \begin{minipage}[b]{1.0\linewidth}
    \centering
    \includegraphics[width=1.0\hsize]{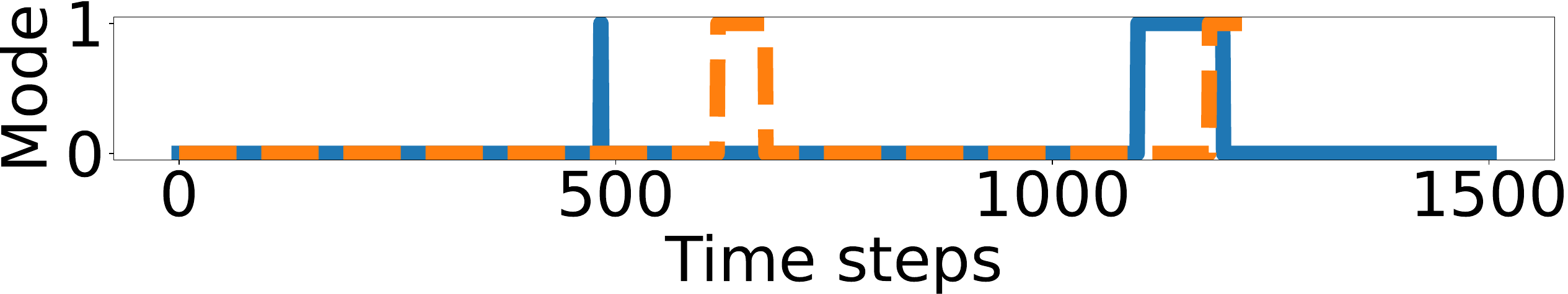}
    \subcaption{BC w/ swtich}
    \end{minipage}

\caption{
Results of mode switching by learned mode-switching policy (blue line) in testing phase: Orange dotted line is demonstrated mode switching. Ours performed mode-switching similar to demonstration, while other methods failed to correctly perform mode-switching.
}
\label{fig:evaluation_simulation_mode}
\end{figure}

\begin{figure}[tb]
    \centering

    \begin{minipage}[b]{0.48\linewidth}
    \centering
    \includegraphics[width=1.0\hsize]{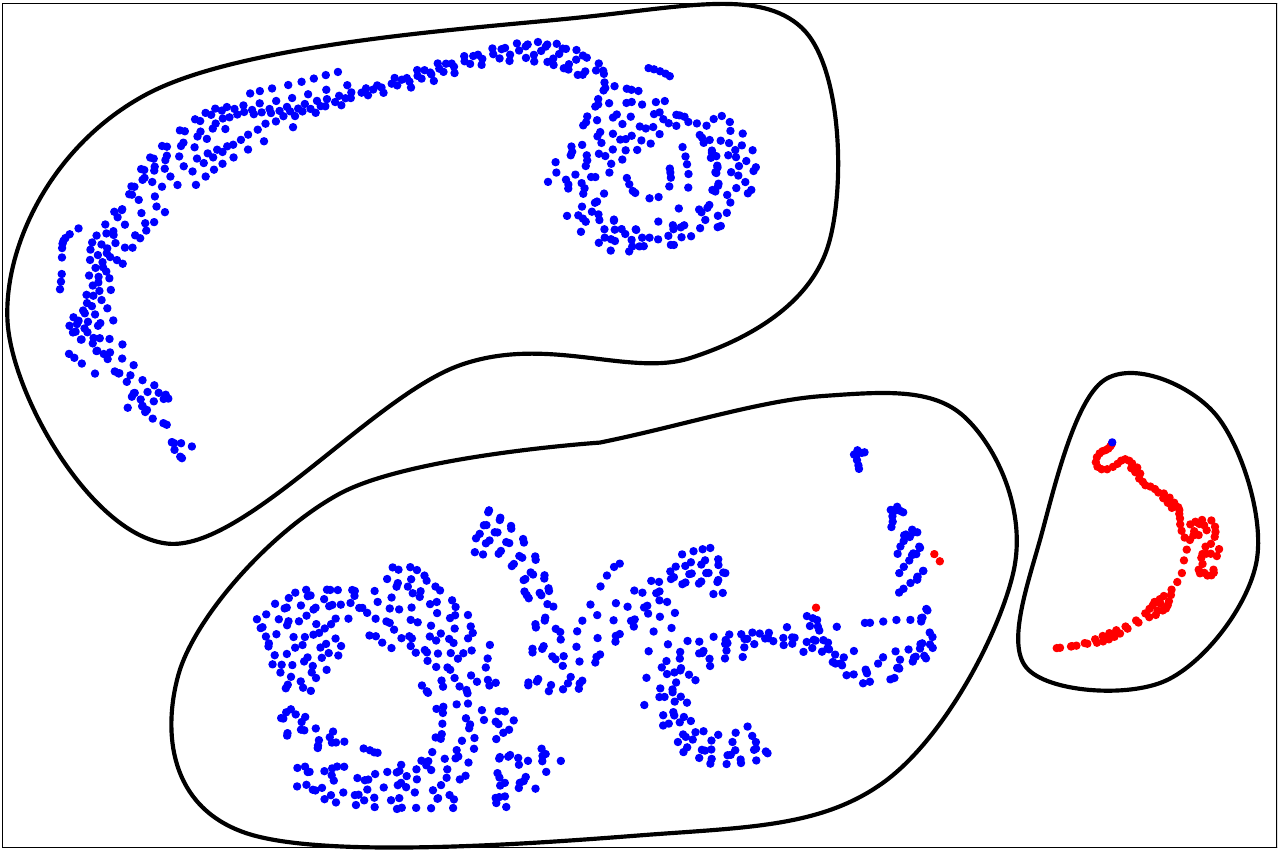}
    \subcaption{Ours}
    \end{minipage}
    \begin{minipage}[b]{0.48\linewidth}
    \centering
    \includegraphics[width=1.0\hsize]{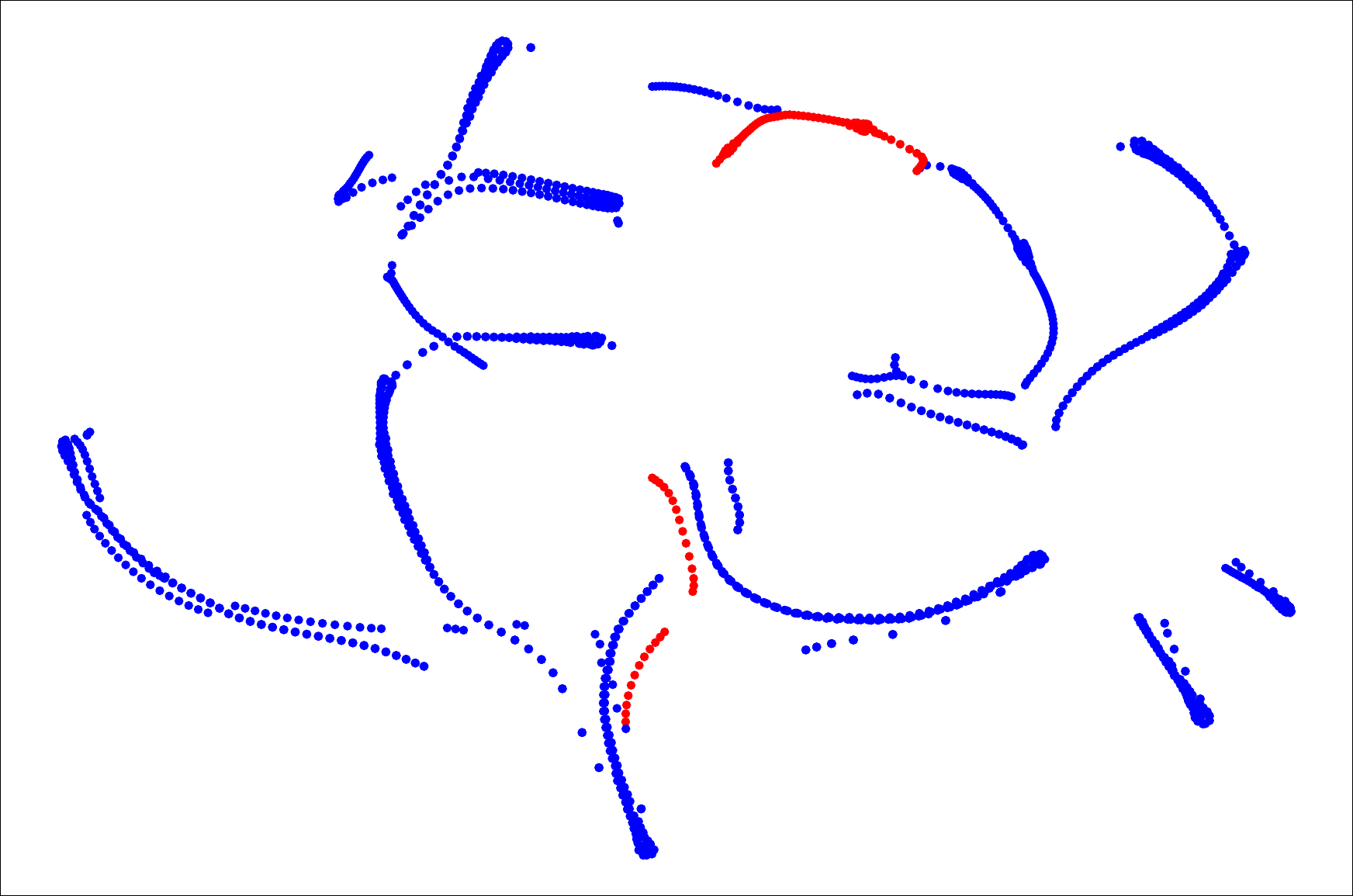}
    \subcaption{BC w/ switch}
    \end{minipage}
    % \begin{minipage}[b]{0.48\linewidth}
    % \centering
    % \includegraphics[width=1.0\hsize]{./figures/evaluation/simulation/Ours_wo_past/t-SNE.pdf}
    % \subcaption{Ours w/o past}
    % \end{minipage}

    % \begin{minipage}[b]{0.48\linewidth}
    % \centering
    % \includegraphics[width=1.0\hsize]{./figures/evaluation/simulation/Ours_wo_reg/t-SNE.pdf}
    % \subcaption{Ours w/o reg}
    % \end{minipage}
    % \begin{minipage}[b]{0.48\linewidth}
    % \centering
    % \includegraphics[width=1.0\hsize]{./figures/evaluation/simulation/BC_w_switch/t-SNE.pdf}
    % \subcaption{BC w/ switch}
    % \end{minipage}
    
    % \begin{minipage}[b]{0.48\linewidth}
    % \centering
    % \includegraphics[width=1.0\hsize]{./figures/evaluation/simulation/BC_w_belief/t-SNE.pdf}
    % \subcaption{BC w/ belief}
    % \end{minipage}
    
\caption{
Visualization of two-dimensional compression of (a) estimated belief states and (b) observations on test trajectory by applying t-SNE: Red and blue points represent either task-oriented action or exploration labels.
}
\label{fig:evaluation_simulation_tsne}
\end{figure}

\tableref{table:evaluation_simulation_performance_ablation} shows another analysis of the effect of future/past steps addressed for belief state regularizations. $k=10$ achieved a better success rate than $k=3$ since the former considers long-term transitions as explicit regularization, which is beneficial for capturing accurate latent states. On the other hand, $k=20$ did not further improve the success rate, and the training time was increased due to the complexity of the model optimization. These results suggest that considering $k=10$ is a reasonable setting for this task. \figref{fig:evaluation_simulation_mode} reflects the benefit of our belief state estimation that contributes to the appropriate mode-switching that resembles the demonstration's mode-switching, in contrast to the comparison methods that failed to appropriately determine the mode-switching timing, causing a task failure. The visualization of a two-dimensional compression of the estimated high-dimensional belief states by applying t-SNE \cite{van2008visualizing} (\figref{fig:evaluation_simulation_tsne}) provides deeper analysis for this aspect. As shown in Ours, the belief states can be separated into three clusters, classified as an exploration forward direction, an exploration backward direction, and a task-oriented action, that is the expected results of separating the states (\figref{fig:evaluation_simulation_environment}). In contrast, the baseline method BC w/ switch failed because it does not have belief state estimation, as shown as multiple task-oriented action clusters. 

%%%%%%%%%%%%%%%%%%%%%%%%%%%%%%%%%%%%%%%%%%%%%%%%%%%%%%%%%%%%%%%%%%%%%%%%%%%%

%%%%%%%%%%%%%%%%%%%%%%%%%%%%%%%%%%%%%%%%%%%%%%%%%%%%%%%%%%%%%%%%%%%%%%%%%%%%
\subsection{Real robot experiment}

\subsubsection{Task setting}

\begin{figure}[tb]
\centering
\includegraphics[width=1.0\hsize] {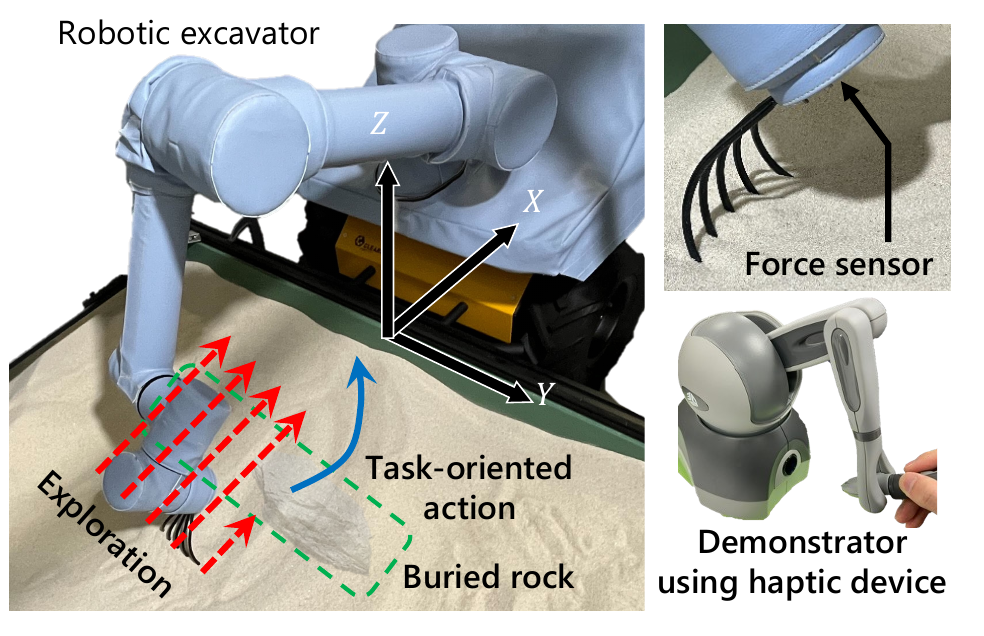}
\caption{
Real robot environment for rock-removal task: Demonstrator initially utilized a pre-defined exploration policy (red dotted arrows). For task-oriented action demonstration (blue arrow), it used a 3D SYSTEMS haptic device with force feedback for robot operation by feeling force interaction between robot's end-effector and sand/rock. 
}
\label{fig:evaluation_realrobot_environment}
\end{figure}

Our experiment performed a rock-removal task that removed a buried rock that cannot be directly observed, in contrast to a previous study\cite{sotiropoulos2020autonomous}. Sotiropoulos et al. \cite{sotiropoulos2020autonomous} conducted a similar rock-removal task, although they assumed that the rock’s position and orientation can be observed using a motion capture system. As shown in \figref{fig:evaluation_realrobot_environment}, our robotic arm imitated an excavator. A claw tool was attached to a Universal Robot UR5e as its end-effector, which dug out the buried rock while moving through the sand. The rock was buried at a random initial position ($X, Y, Z$) (area indicated by the green dotted line). The demonstrator ended the task when the rock was completely removed from the sand and brought to the surface.

\subsubsection{Demonstration setting}

Three human subjects with basic knowledge of robot control practiced the following operation. For a demonstration with switching, the demonstrators collected information about the rock using the automatic exploration policy. During the exploration policy, the demonstrators monitored the force/torque sensor values and the robot's behavior, while pausing the manual operations. They switched from exploration to the task-oriented action policy based on his/her estimated latent rock belief state by pressing a button on the haptic device. In contrast, the demonstration without a switching structure requires arbitrary exploration for demonstrators, as in \figref{fig:introduction_problem}(a). During task-oriented actions, scaled force feedback is provided to the haptic device based on the robotic arm's force torque sensor values attached to the robot's wrist. The demonstrator performed robot operation while perceiving the feedback that is associated with contact with the soil, the rock, the robot's end-effector, and occasionally utilizing visual and sound information. The observations used for the policy learning are the position $(X, Y, Z, \theta)$ of the robot's end-effector, the end-effector's movement speed $(\dot{X},\dot{Y},\dot{Z},\dot{\theta})$, and the 6-D force torque sensor values. The predicted action is the end-effector's movement speed $(\dot{X},\dot{Y},\dot{Z},\dot{\theta})$.

\subsubsection{Learning setting}

The models were trained using ten demonstration trajectories, and each method was tested ten times. Other settings are identical as in the simulation \sref{sec:simulation}.

\subsubsection{Result}

\begin{table*}[tb]
\centering
\caption{
Results of learned policies' success rate, and mode prediction (pred.) accuracy (acc.), and action prediction (pred.) accuracy (acc.) in the real robot experiment. Please note that each value of action prediction loss has [$\times 10^{-4}$], which is omitted for space limitation. Ours achieved the best performance, the highest mode prediction accuracy, and the lowest action prediction loss against comparison methods.
}
\label{table:evaluation_realrobot_result}
\begin{tabular}{l | c c c}
\hline
\diagbox[width=2.7cm, height=\line]{\ }{\ } & Success rate [\%] & Mode pred. acc. [\%] & Action pred. loss \\
\hline
\hline
Ours & $\mathbf{80 \pm 14}$ & $\mathbf{99.5 \pm 0.23}$ & $\mathbf{0.81 \pm 0.07}$ \\
Ours w/o reg  & $35 \pm 35$ & $91.2 \pm 5.35$ & $3.08 \pm 0.09$ \\
BC w/ switch & $20 \pm 14$ & $85.0 \pm 2.09$ & $2.60 \pm 0.05$ \\
BC w/ belief & $15 \pm 7$ & $--$ & $6.68 \pm 3.66$ \\
BC & $5 \pm 7$ & $--$ & $40.6 \pm 8.42$ \\
\hline
\end{tabular}
\end{table*}

\begin{figure}[tb]
\centering
\includegraphics[width=1.0\hsize] {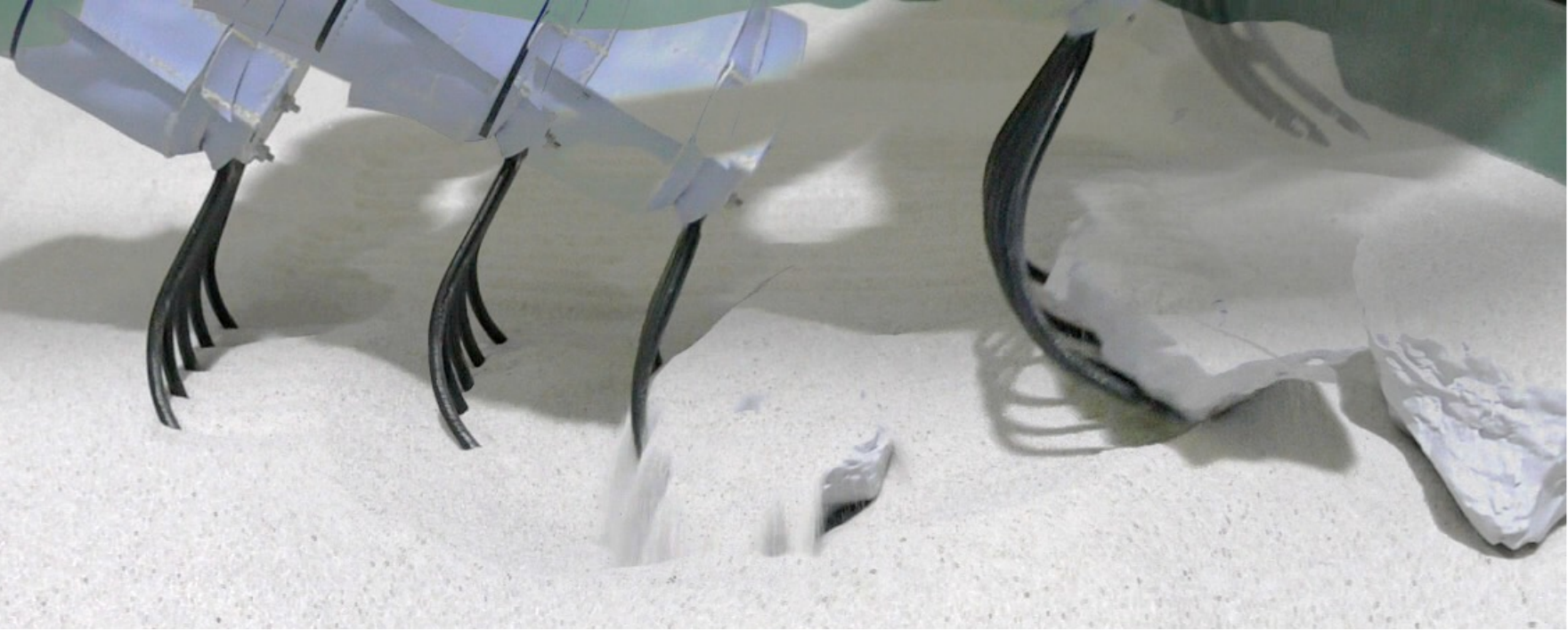}
\caption{
Time-series snapshots of successful rock removal motion by Ours.
}
\label{fig:evaluation_realrobot_taskexample}
\end{figure}

The learned policies' test performances (success rate) and mode and action prediction accuracies are shown in \tableref{table:evaluation_realrobot_result}. Ours achieved the best performance, the highest mode prediction accuracy, and the lowest action prediction loss. In contrast, Ours w/o reg, had a lower success rate due to the absence of belief state regularizations. \figref{fig:evaluation_realrobot_taskexample} shows the successful rock removal motion of Ours. The baseline methods (BC w/ switch, BC w/ belief, and BC) also had poorer performances due to the absence of mode-switching (causing inconsistent action demonstration) and belief state estimation.

\begin{table}[htbp]
\centering
\caption{
Result of user study with three subjects who evaluated cognitive load in demonstration: Based on NASA TLX, subjects evaluated mental demand and frustration on following five levels: 1) very low, 2) low, 3) medium, 4) high, and 5) very high \cite{hart2006nasa}. For mental demand, a significant difference by t-test ($p < 0.05$) was observed between demonstration with and without switching. Right column shows number of failures (emergency stops) caused by strong collision between robot's claw and rock.
}
\label{table:evaluation_realrobot_demonstration_burden}
\begin{tabular}{l | c | c | c}
\hline
Method & Mental demand & Frustration & Failures \\
\hline
\hline
\textbf{Demo w/ switch}     & $\mathbf{2.3 \pm 0.6}$    & $\mathbf{2.7 \pm 0.6}$    & $\mathbf{0.7 \pm 0.6}$ \\
Demo w/o switch             & $4.0 \pm 0.0$             & $3.7 \pm 0.6$             & $2.7 \pm 1.5$\\
\hline
\end{tabular}
\end{table}

\tableref{table:evaluation_realrobot_demonstration_burden} shows the cognitive load (mental demand and frustration) between a demonstration with switching, which uses the automatic exploration policy, and a demonstration without switching, which entrusts the entire operation to the demonstrator's decision-making. In both mental demand and frustration, which are the immediate mental demand and cumulative mental demand over a series of demonstrations, Demo w/ switch reduced the cognitive load more than Demo w/o switch. The mental demand shows an especially significant difference by t-test ($p < 0.05$). These results are reflected in the number of failures; Demo w/ switch achieved fewer failures (emergency stops) than Demo w/o switch, owing to a smaller cognitive load. 
%%%%%%%%%%%%%%%%%%%%%%%%%%%%%%%%%%%%%%%%%%%%%%%%%%%%%%%%%%%%%%%%%%%%%%%%%%%%

%%%%%%%%%%%%%%%%%%%%%%%%%%%%%%%%%%%%%%%%%%%%%%%%%%%%%%%%%%%%%%%%%%%%%%%%%%%%
\section{Discussion}

The following are the two main advantages of our proposed method:
\begin{enumerate*}
\item \textbf{High performance and prediction accuracy}: the proposed method achieved higher performance and higher mode and action prediction accuracies than the comparison methods, due to a belief state estimation with future and past regularization and consistent action demonstration using the pre-designed exploration policy.
\item \textbf{Less cognitive load in demonstration}: it reduced the cognitive load on the demonstrators with a pre-designed exploration policy.
\end{enumerate*}

Our future works will address the following limitations:
\begin{enumerate*}
\item This study assumes that the demonstrator collects additional binary mode labels. Although extending the number of modes is possible, the cognitive load of mode switching increases accordingly. For this aspect, the unsupervised method could be useful for extracting a policy-segmented structure \cite{tsai2019unsupervised}.
\item The exploration policies used in the experiment are empirically designed in a feed-forward manner. For more practical applications, we want to change the exploration policy to a state-aware or a parameterized policy and optimize the parameters.
\item Although the most basic imitation learning approach, Behavior Cloning was used to validate our proposed method in this study, validation against other imitation learning methods (e.g., DAgger Family \cite{ross2011reduction, hoque2022thriftydagger}) is one of the future investigations.
\end{enumerate*}
%%%%%%%%%%%%%%%%%%%%%%%%%%%%%%%%%%%%%%%%%%%%%%%%%%%%%%%%%%%%%%%%%%%%%%%%%%%%

%%%%%%%%%%%%%%%%%%%%%%%%%%%%%%%%%%%%%%%%%%%%%%%%%%%%%%%%%%%%%%%%%%%%%%%%%%%%
\section{Conclusion}

This paper proposed a novel imitation learning framework for nonprehensile manipulation tasks of invisible objects, that switches between a pre-designed exploration policy and a task-oriented action policy using a belief state estimator trained with regularization using future and past information. In a simulation and real robot nonprehensile manipulation tasks of invisible objects, we confirmed that the proposed method achieved the best mode and action prediction accuracies and task performance and reduced the cognitive load in our demonstration.
%%%%%%%%%%%%%%%%%%%%%%%%%%%%%%%%%%%%%%%%%%%%%%%%%%%%%%%%%%%%%%%%%%%%%%%%%%%%

\section*{Acknowledgments}
This work is supported by JST [Moonshot Research and Development], Grant Number [JPMJMS2032].

\bibliographystyle{elsarticle-num}
\bibliography{paper}

\end{document}